\documentclass[lettersize,journal]{IEEEtran}
\usepackage{amsmath,amsfonts}
\usepackage{algorithmic}
\usepackage{algorithm}
\usepackage{array}
\usepackage{multirow}
\usepackage[caption=false,font=normalsize,labelfont=sf,textfont=sf]{subfig}
\usepackage{textcomp}
\usepackage{stfloats}
\usepackage{url}
\usepackage{verbatim}
\usepackage{graphicx}
\usepackage{cite}
\usepackage{balance}

\begin{document}
\title{GeoGS-SLAM: Geometry-Only Gaussian Splatting for Dense Monocular SLAM}

\author{Lipu Zhou$^{*1}$, Yaoyun Kang$^{1}$, Junxiang Pang$^{1}$, Shengkai Sun$^{1}$, Tingting Bao$^{1}$, Kehan Wang$^{1}$%
\thanks{*Corresponding author} 
\thanks{$^{1}$The authors are with the School of Instrument Science and Opto-electronics Engineering, Beihang University, Beijing 100191, China (e-mail: \tt\small \{zhoulipu, KangYY, pangjunxiang, 22373368, btt2022, 25171012\}@buaa.edu.cn)}}


\maketitle

\begin{abstract}
Dense visual Simultaneous Localization and Mapping (SLAM) is a fundamental problem in robotics. Recent advances in 3D Gaussian Splatting (3DGS) have demonstrated its potential for dense SLAM. Existing 3DGS frameworks focus on both appearance and geometry modeling. However, scene geometry is typically more critical for SLAM than novel view synthesis because downstream robotic tasks, such as navigation and obstacle avoidance, rely primarily on accurate spatial geometry rather than photorealistic rendering. This observation raises a natural question: Is it feasible for 3DGS to perform 3D reconstruction without scene appearance modeling? Motivated by this, we propose Geometry-only Gaussian Splatting (GeoGS), which directly reconstructs scene geometry, and further present GeoGS-SLAM, a dense visual SLAM system built upon this representation. Specifically, GeoGS retains only spatial parameters (position, rotation, scale, and opacity) to reduce the number of per-primitive parameters by over 80\%. In contrast to existing 3DGS methods, GeoGS focuses solely on geometric reconstruction, which significantly reduces the number of Gaussian primitives, accelerates geometric convergence, and enhances robustness to illumination variations. In addition, we present an effective training framework that optimizes the Gaussian primitives via single-view and multi-view geometric and photometric supervision, and speeds up geometry convergence with a local-plane driven initialization that better aligns primitives with local structures. Furthermore, we introduce a map update strategy for loop closure or global Bundle Adjustment (BA) that globally transforms the Gaussian map to align it with the corrected pose estimates while preserving local structural coherence, thereby preventing map tearing caused by inconsistent per-viewpoint pose corrections in existing methods. Extensive experiments on synthetic and real-world benchmarks demonstrate that our method outperforms state-of-the-art methods in terms of online mapping efficiency and geometric reconstruction quality. Our code will be made available.
\end{abstract}


\begin{figure}[t]
\centering
\includegraphics[width=0.49\textwidth]{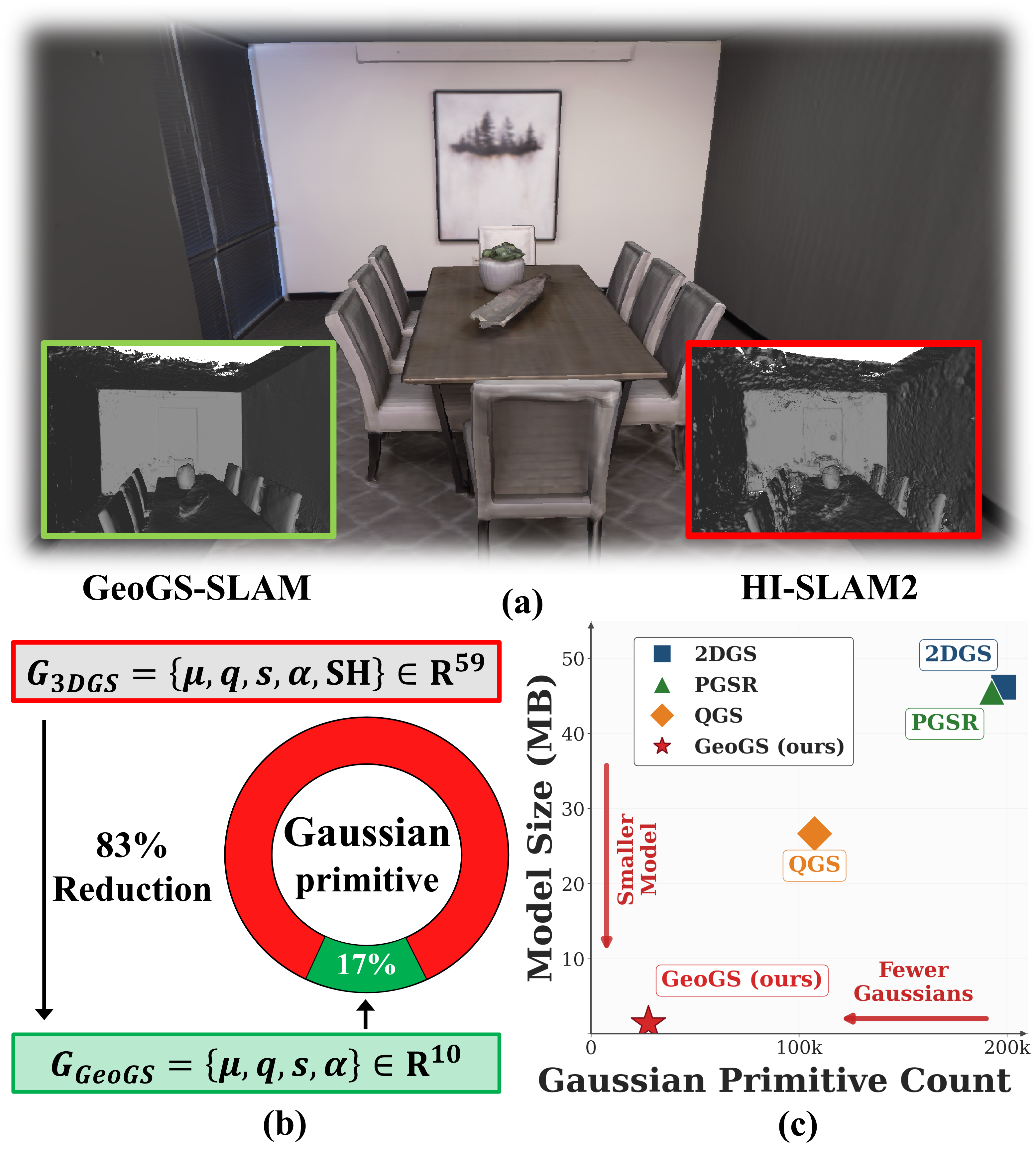}
\caption{\textbf{Illustration of the characteristics of GeoGS and the performance of GeoGS-SLAM.} (a) Qualitative comparison of geometric reconstruction. GeoGS-SLAM recovers much cleaner scene geometry compared to the baseline method. (b) Per-primitive parameter efficiency. By retaining only the geometr-related parameters ($\mathbf \mu$,$\mathbf q$,$\mathbf s$,$\alpha$), GeoGS remains only 17\% of the parameters required by a standard 3DGS primitive. (c) Model efficiency of the reconstructed Gaussian map. In addition to the reduction of per-primitive parameters, GeoGS represents the scene with substantially fewer Gaussians, requiring only 28k Gaussian primitives, compared with 198k for 2DGS, 193k for PGSR, and 107k for QGS.}
\label{geogs}
\end{figure}

\section{Introduction}
Dense visual Simultaneous Localization and Mapping (SLAM), which jointly estimates camera poses and reconstructs a dense 3D map of the environment, is a fundamental problem in robotics\cite{carlone2025slam}. Recently, 3D Gaussian Splatting (3DGS)\cite{kerbl20233d} introduces a new paradigm for 3D reconstruction. Through explicit Gaussian-based scene representation and differentiable rasterization, 3DGS has demonstrated remarkable performance in novel view synthesis (NVS) and has emerged as a promising approach for dense visual SLAM\cite{deng2025best}.

Following the success of 3DGS and its variants in NVS\cite{yu2024mip,fang2024mini,mallick2024taming,cheng2024gaussianpro}, a growing body of work has shifted attention toward geometry reconstruction. Existing 3DGS-based visual reconstruction methods attempt to improve geometry quality by introducing explicit geometric regularizations that encourage Gaussians to lie near scene surfaces rather than floating in free space\cite{turkulainen2025dn, chen2024pgsr} or by adapting Gaussian parameterizations to better align with scene surfaces\cite{huang20242d, zhang2025quadratic}. Despite these efforts, each Gaussian primitive is inherently parameterized by both color and spatial attributes, representing appearance and geometry together. However, scene geometry is more critical for SLAM than view synthesis. Downstream applications such as navigation and obstacle avoidance rely primarily on accurate spatial geometry rather than on how realistically the novel view is rendered. Under this coupled design, floating artifacts that are geometrically imprecise but help reduce photometric error tend to be preserved rather than corrected\cite{shen2025evolving}. In addition, the photometric objective itself demands a larger number of per-primitive parameters to model view-dependent appearance. Consequently, geometry optimization becomes less effective under the limited per-keyframe update budget of online dense mapping.

Beyond the appearance-geometry coupled formulation, 3DGS‑SLAM faces an additional challenge when the map must be globally corrected. After loop closure or global Bundle Adjustment (BA) resolves the accumulated drift, the corrected camera trajectory becomes inconsistent with the existing Gaussian map. Retraining the map from revisited keyframes is computationally prohibitive for real-time performance. To alleviate these discrepancies efficiently, existing methods typically associate each Gaussian primitive with an anchor keyframe and propagate the viewpoint correction directly to the parameters of the associated Gaussians. While this per-viewpoint propagation maintains map consistency after trajectory correction without full retraining, it disregards spatial coherence: neighboring Gaussians that belong to the same surface but are observed from different keyframes can receive different, sometimes conflicting, transformations. This pulls adjacent primitives apart and introduces tearing and floating artifacts, thereby degrading the structural coherence of the reconstructed geometry.

To address these challenges, we first propose Geometry-only Gaussian Splatting (GeoGS) to reconstruct scene geometry directly. As illustrated in Fig.~\ref{geogs}, unlike existing 3DGS-based methods that encode scene appearance using Spherical Harmonics (SH) coefficients, GeoGS directly represents scene geometry using only geometry-related parameters, including position $\mu$, rotation $q$, scale $s$, and opacity $\alpha$, without appearance modeling. This design reduces the number of per-primitive parameters by over 80\% and requires fewer primitives to represent scene geometry, enabling faster geometric convergence while improving robustness to illumination variations. Building upon GeoGS, we further develop GeoGS-SLAM for dense monocular SLAM.

Moreover, we present a training framework to optimize Gaussian primitives, as conventional color rendering supervision is no longer available in our GeoGS settings. The framework incorporates single-view and multi-view geometric and photometric supervision for Gaussian optimization, together with a local-plane driven initialization strategy that aligns Gaussians with locally estimated planar structures via Principal Component Analysis (PCA) to accelerate geometric convergence under constrained online optimization budgets.

Furthermore, to resolve the map tearing and floating artifacts caused by map corrections after loop closure or global Bundle Adjustment (BA), we introduce a coherent map update strategy. Instead of propagating viewpoint corrections independently to associated Gaussians, we model the correction as a unified Sim(3) transformation applied to regions of revisited Gaussians. By treating map deformation as a holistic geometric adjustment rather than a collection of per-viewpoint displacements, this strategy preserves local structural coherence and substantially reduces the artifacts.

In summary, our main contributions are as follows:
\begin{enumerate}
  \item We propose GeoGS, which reconstructs scene geometry using only geometry-related parameters (e.g., position, rotation, scale, and opacity), reducing per-primitive parameters by over 80\%. Without appearance modeling, GeoGS requires fewer primitives and is inherently less sensitive to lighting variations. Together, these benefits enable faster geometric convergence.

  \item We present a training framework for GeoGS optimization, consisting of single-view and multi-view geometric and photometric supervision for optimization without conventional color rendering supervision, and a local-plane driven initialization strategy that aligns primitives with local geometric structures to further accelerate convergence in geometry.
  \item We introduce a novel map update strategy after loop closure or global Bundle Adjustment (BA) that globally aligns the Gaussian map with the corrected poses while maintaining local geometric consistency. In contrast to existing methods propagating per-viewpoint pose corrections, which pull adjacent Gaussians apart and cause map tearing, our approach estimates a unified Sim(3) transformation for all revisited Gaussians.
  \item Extensive experiments on synthetic and real-world datasets demonstrate that GeoGS-SLAM outperforms state-of-the-art methods. In particular, it improves mean accuracy and mean completeness by 25\% and 5\%, respectively, on the Replica dataset, and reduces Chamfer Distance by 30\% on the ScanNet++ dataset.
\end{enumerate}

\section{Related Works}
\subsection{3DGS-based Visual Reconstruction}

3D Gaussian Splatting (3DGS)\cite{kerbl20233d} introduces a differentiable scene representation together with CUDA-accelerated rasterization, enabling efficient optimization of Gaussian primitives. Building upon 3DGS, early methods\cite{fang2024mini,yu2024mip,cheng2024gaussianpro,lu2024scaffold,radl2024stopthepop} focus on novel view synthesis (NVS) and aim to improve rendering quality. The success of 3DGS in NVS also motivates growing interest in geometry reconstruction and surface modeling through Gaussian representations.

To improve geometric accuracy, explicit geometric constraints have been incorporated into Gaussian optimization. SuGaR\cite{guedon2024sugar} regularizes Gaussian primitives to remain close to an extracted surface and enables mesh extraction. GOF\cite{yu2024gaussian} models Gaussian opacity as a continuous field and directly recovers geometry through level-set extraction. RaDe-GS\cite{zhang2026rade} introduces rasterized depth and normal rendering for general 3D Gaussian primitives. GSDF\cite{yu2024gsdf} combines Gaussian primitives with signed distance fields to improve geometric reconstruction quality. PGSR\cite{chen2024pgsr} further introduces multi-view geometric and photometric constraints to enhance geometric consistency and surface reconstruction quality.

Beyond introducing geometric constraints during optimization, researchers also enhance reconstruction quality through improved primitive designs that better align with scene surfaces. 2DGS\cite{huang20242d} replaces volumetric ellipsoids with oriented planar disks and improves per-primitive depth accuracy through ray-splat intersection. MP-GS\cite{qu2025mixed} represents the scene with multiple types of primitives, enabling a more flexible geometric representation of scene structures. QGS\cite{zhang2025quadratic} extends Gaussian primitives with second-order geometric fitting capability for curved surfaces. GGGS\cite{zhang2026geometry} further explores stochastic solid representations for more expressive geometric modeling.

\subsection{Dense Visual SLAM}

Classical dense visual SLAM systems focus on jointly estimating camera motion and reconstructing scene geometry. DTAM\cite{newcombe2011dtam} formulates dense monocular reconstruction as variational optimization on depth maps. KinectFusion\cite{newcombe2011kinectfusion} introduces online TSDF fusion for incremental dense reconstruction from RGB-D input, while ElasticFusion\cite{whelan2016elasticfusion} extends surfel-based mapping with non-rigid deformation to improve global consistency. BundleFusion\cite{dai2017bundlefusion} further combines global pose optimization with online surface fusion for globally consistent dense reconstruction. LSD-SLAM\cite{engel2014lsd} proposes a semi-dense map representation that enables real-time monocular SLAM. With the development of deep learning, CNN-SLAM\cite{tateno2017cnn} utilizes convolutional neural networks (CNN) to predict the dense map. CodeSLAM\cite{bloesch2018codeslam} employs an autoencoder for depth prediction. Building upon RAFT\cite{teed2020raft}, DROID-SLAM\cite{teed2021droid} further achieves a robust dense visual SLAM and has been widely adopted for dense mapping systems.

The emergence of NeRF\cite{mildenhall2021nerf} has introduced a new paradigm for dense visual SLAM by enabling photorealistic scene rendering. Early systems such as iMAP\cite{sucar2021imap} and NICE-SLAM\cite{zhu2022nice} integrate implicit neural representations into a SLAM framework. NICER-SLAM\cite{zhu2024nicer} further extends this paradigm to RGB-only input. To improve representation efficiency, ESLAM\cite{johari2023eslam} employs axis-aligned feature planes instead of fully implicit scene models. Building upon DROID-SLAM and Instant-NGP\cite{muller2022instant}, GO-SLAM\cite{zhang2023go} adopts multi-resolution hash representation for faster reconstruction, while HI-SLAM\cite{zhang2023hi} and GLORIE-SLAM\cite{zhang2024glorie} further incorporate monocular depth priors to enhance geometric consistency. Despite achieving impressive rendering quality in online mapping, these methods typically require computationally intensive neural optimization and rely on large voxel grids or multilayer perceptrons (MLP) to represent the scene, limiting their efficiency and scalability for real-time dense SLAM.

With recent advance in feed-forward reconstruction methods such as MASt3R\cite{leroy2024grounding}, VGGT\cite{wang2025vggt}, and Depth-Anything v3\cite{lin2025depth}, an increasing number of SLAM systems build upon these pretrained reconstruction models. MASt3R-SLAM leverages the two-view 3D reconstruction priors extracted from MASt3R for tracking and mapping. VGGT-SLAM\cite{maggio2026vggt} utilizes submap alignment for multi-view reconstruction. VGGT-SLAM 2.0\cite{maggio2026vggt} removes high-dimensional 15-degree-of-freedom drift and planar degeneracy, achieving higher accuracy in both tracking and mapping. VGGT-Long\cite{deng2025vggt} introduces a chunk-based processing strategy and extends VGGT to long sequences. VGGT-Motion\cite{vggt-motion} proposes a motion-aware submap construction strategy and mitigates drift in long sequences. However, the substantial computational overhead of feed-forward models generally restricts these methods to processing a limited set of selected keyframes, thereby producing temporally sparse camera trajectories\cite{hyvggt_vo_2026}.

\subsection{3DGS-based Visual SLAM}
Recently, 3D Gaussian Splatting (3DGS) has emerged as an attractive explicit representation for dense visual SLAM. Early GS-based SLAM systems, including GS-SLAM\cite{yan2024gs}, SplaTAM\cite{keetha2024splatam}, and MonoGS\cite{matsuki2024gaussian}, directly incorporate Gaussian primitives as explicit dense maps into the SLAM pipeline, demonstrating that 3DGS can achieve promising reconstruction quality in SLAM\cite{wang2026towards}. Gaussian-SLAM\cite{yugay2023gaussian} utilizes submap management to ensure photorealistic rendering in large environments. Photo-SLAM\cite{huang2024photo} introduces hyper primitives and improves the rendering quality. RTG-SLAM\cite{peng2024rtg} proposes a map management strategy that forces Gaussian primitives to be opaque or transparent, using a single opaque Gaussian to fit a local surface without the need for multiple overlapping Gaussians. CG-SLAM\cite{hu2024cg} introduces an uncertainty-aware 3D Gaussian field and achieves high consistency and geometric stability. FGS-SLAM\cite{xu2025fgs} introduces frequency domain analysis to realize high-quality reconstruction. MG-SLAM\cite{liu2025mg} introduces the Manhattan world assumptions to enhance structural coherence. SEGS-SLAM\cite{tianciwenSEGSSLAM2025} introduces Scaffold-GS\cite{lu2024scaffold} into the SLAM system and achieves improved rendering quality. To further improve the mapping quality in terms of geometry reconstruction, MGS-SLAM\cite{zhu2024mgs} leverages Multi-view Stereo (MVS) networks to obtain dense depth maps. Subsequent works further integrate stronger frontend priors into Gaussian-based SLAM systems. Incorporating DROID-SLAM as the frontend, Splat-SLAM\cite{sandstrom2025splat} and DROID-Splat\cite{homeyer2025droid} utilize monocular depth estimates as pseudo RGB-D input. HI-SLAM2\cite{zhang2025hi} further incorporates monocular normal priors to improve the quality of geometric reconstruction. Building upon DSO\cite{engel2017direct}, MGSO\cite{hu2025mgso} employs 3DGS as the scene representation and runs scene reconstruction in parallel with camera tracking. GSO-SLAM\cite{yeon2026gso} proposes a tightly coupled system, enabling joint optimization of tracking and mapping. GaussianFlow SLAM\cite{seo2026gaussianflow} aligns the projected motion of Gaussians with the optical flow to regularize both tracking and mapping. S3PO-GS\cite{cheng2025outdoor} and ARTDECO\cite{li2025artdeco} also introduce geometric priors from MASt3R to enhance camera tracking. Flash-Mono\cite{zhang2026flash} further introduces feed-forward Gaussian prediction to accelerate the mapping process in GS-SLAM.

With these advances, pose correction, such as loop closure and global bundle adjustment, poses another challenge for GS-based SLAM. It requires not only correcting accumulated pose and scale drift, but also updating the explicit Gaussian map to maintain consistency with the corrected trajectory.

Several recent studies address this issue through different map update strategies after loop closure. VINGS-Mono\cite{wu2025vings} restores consistency by revisiting loop-affected regions and retraining them after trajectory correction. Although such reconstruction-based refinement resolves the inconsistency between the corrected trajectory and the existing Gaussian map, the additional optimization process introduces extra latency in map updating. Some GS-based SLAM systems adopt a direct Gaussian correction after trajectory optimization. GLC-SLAM\cite{xu2024glc}, VIGS-SLAM\cite{zhu2025vigs} and 2DGS-SLAM\cite{zhong2026globally} associate Gaussian primitives with keyframes from which the Gaussians are initialized, and update the map by propagating the corresponding $\mathrm{SE}(3)$ or $\mathrm{Sim}(3)$ pose correction to the associated Gaussians. This rapid update strategy avoids explicit retraining and enables immediate map correction after loop closure. However, this correction strategy has a fundamental limitation: neighboring Gaussians associated with different keyframes may receive different transformations after global trajectory optimization. As a result, geometrically adjacent primitives on the same surface may become spatially inconsistent, leading to local structural artifacts such as surface tearing or layered geometry.

In this work, we propose GeoGS-SLAM, a dense monocular SLAM system that introduces GeoGS for efficient geometry reconstruction and preserves global map consistency through coherent Gaussian map updates after trajectory correction.

\begin{figure*}[t]
\centering
\includegraphics[width=0.98\textwidth]{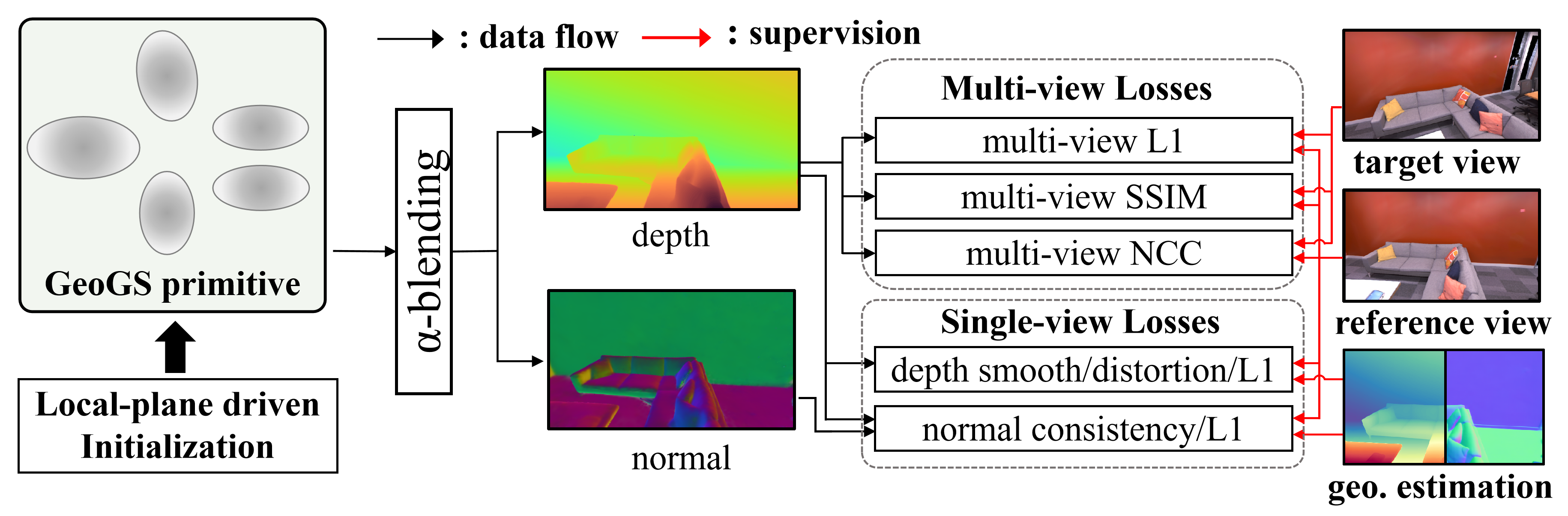}
\caption{Overview of the GeoGS mapping framework. GeoGS primitives are initialized using a local-plane driven strategy and optimized through single-view and multi-view losses. The single-view losses supervise the rendered geometry through depth smoothness, depth distortion, normal consistency, and geometric priors from frontend. For multi-view supervision, the rendered depth map of the reference view is used for 3D reprojection to obtain valid correspondences in the target view, over which L1, SSIM, and NCC losses are applied to provide corss-view photometric supervision for geometry optimization.}
\label{fig:geogs_mapping}
\end{figure*}

\section{Preliminaries of 3D Gaussian Splatting}
3D Gaussian Splatting (3DGS) represents a scene as a collection of anisotropic Gaussian primitives $\{\mathcal{G}_i\}$. Specifically, each Gaussian is parameterized by its center position $\mathbf{\mu}_i\in\mathbb{R}^{3}$, rotation $\mathbf{q}_i\in\mathbb{R}^4$, scale $\mathbf{s}_i\in\mathbb{R}^3$, opacity $\alpha_i\in\mathbb{R}$, and spherical harmonics $\mathbf{SH}_i\in\mathbb{R}^{48}$. The spatial distribution of a Gaussian is defined by a 3D Gaussian function:

\begin{equation}
\mathcal{G}_i(x|\boldsymbol{\mu}_i,\boldsymbol{\Sigma}_i) = e^{-\frac{1}{2}(x-\boldsymbol{\mu}_i)^\mathbf{T}\boldsymbol{\Sigma}_i^{-1}(x-\boldsymbol{\mu}_i)},
\label{eq:gaussian_distribution}
\end{equation}
where the covariance matrix $\boldsymbol{\Sigma}_i$ can be decomposed as:

\begin{equation}
\mathbf{\Sigma}_i = \mathbf{R}(\mathbf{q}_i)\,\mathrm{diag}(\mathbf{s}_i)^2\,\mathbf{R}(\mathbf{q}_i)^\top.
\label{eq:gaussian covariance}
\end{equation}

To render an image, the 3D Gaussian is transformed into the camera coordinates with world-to-camera transform matrix $\mathbf{W}$ and projected onto the image plane via a local affine transformation $\mathbf{J}$\cite{zwicker2002ewa}:

\begin{equation}
\mathbf{\Sigma}_i^{'} = \mathbf{J}\mathbf{W}\mathbf{\Sigma}_i\mathbf{W}^\top\mathbf{J}^\top.
\label{eq:gaussian projection}
\end{equation}

By skipping the third row and column of $\mathbf{\Sigma}_i^{'}$, a 2D Gaussian $\mathcal{G}_i^{2D}$ is obtained with a covariance matrix $\mathbf{\Sigma}_i^{2D}$. Then, 3DGS employs alpha blending to integrate alpha-weighted appearance from front to back:
\begin{equation}
\mathbf{c}(x)=\sum_{i=1}^{N}\omega_ic_i,
\label{eq:color rendering}
\end{equation}
where $\mathbf{c}(x)$ represents the rendered color at pixel $x$, $c_i$ represents the color of the i-th Gaussian computed via $\mathbf{SH}_i$, and $\omega_i$ is the contribution of the i-th Gaussian and is computed by:
\begin{equation}
\omega_i=\alpha_i\mathcal{G}_i^{2D}(x)\prod_{j=1}^{i-1}(1-\alpha_j\mathcal{G}_j^{2D}(x)).
\label{eq:weight contribution}
\end{equation}

Typically, given a set of RGB images $\{\mathbf{I_i}\}$ of a scene, the Gaussian primitives are optimized using the color rendering photometric loss via L1 and structural similarity (SSIM) terms:
\begin{equation}
\mathcal{L}=\lambda_1\mathcal{L}_1(\mathbf{\hat{I}}_i,\mathbf{I}_i) + \lambda_{ssim}\mathcal{L}_{ssim}(\mathbf{\hat{I}}_i,\mathbf{I}_i),
\label{eq:color rendering loss}
\end{equation}
where $\mathbf{\hat{I}}_i$ is the rendered RGB image.

\section{GeoGS Mapping}
\label{sec:geogs_mapping}
As shown in Fig.~\ref{fig:geogs_mapping}, this section describes how GeoGS represents and reconstructs scene geometry without color modeling.

\begin{figure}[t]
\centering
\includegraphics[width=0.49\textwidth]{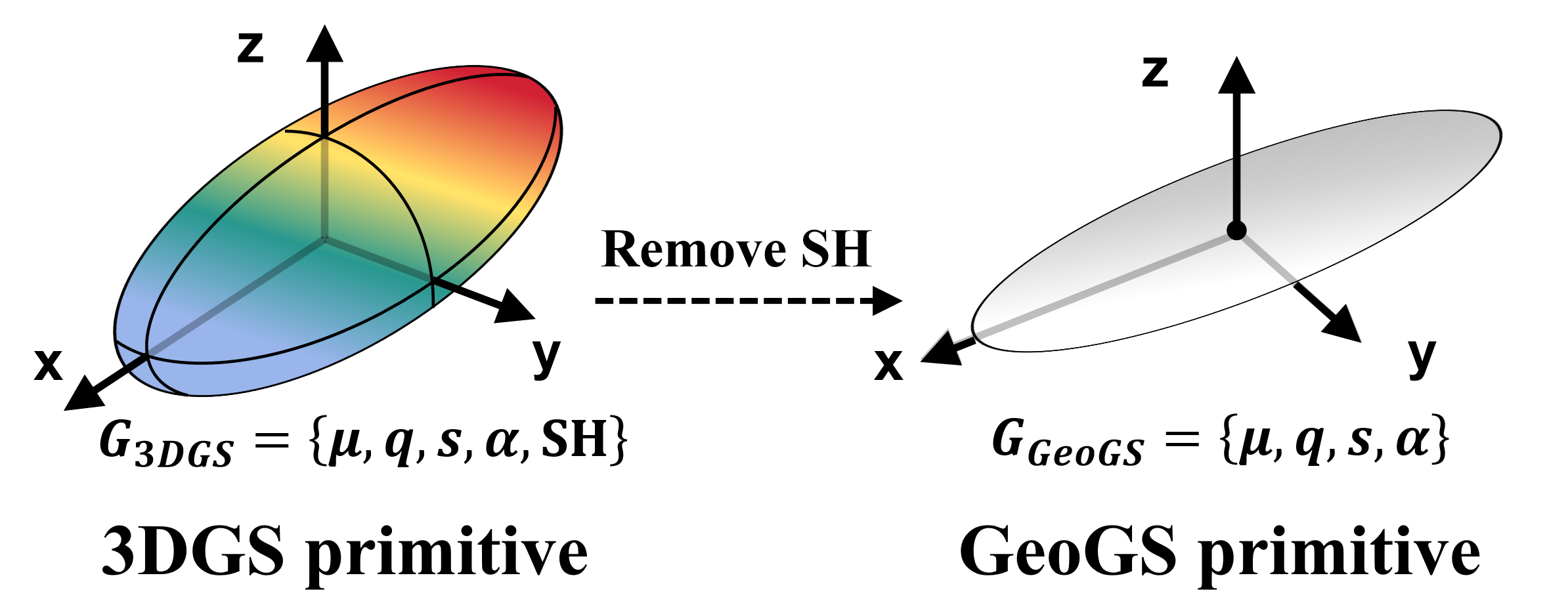}
\caption{Comparison between a standard 3DGS primitive and the proposed GeoGS primitive. GeoGS removes the Spherical Harmonics (SH) and retains only geometry-related parameters for direct scene geometry reconstruction without color modeling. }
\label{fig:geogs_primitive}
\end{figure}

\subsection{GeoGS primitive}
Building upon the Gaussian Splatting representation, GeoGS retains only geometry-related parameters:
\begin{equation}
\mathcal{G}_i=\{\mathbf{\mu}_i,\mathbf{s}_i,\mathbf{q}_i,\mathbf{\alpha}_i\},
\label{eq:geogs_parameter}
\end{equation}
where $\mathbf{\mu}_i\in\mathbb{R}^3$, $\mathbf{s}_i\in\mathbb{R}^2$, $\mathbf{q}_i\in\mathbb{R}^4$, $\mathbf{\alpha}_i\in\mathbb{R}$ denote the position, scale, rotation, and opacity, respectively.

As illustrated in Fig.~\ref{fig:geogs_primitive}, unlike existing 3DGS-based Gaussian representations that encode appearance through Spherical Harmonics, GeoGS removes all appearance-related parameters and reduces the number of parameters per primitive by more than 80\% while preserving the differentiable rasterization process of Gaussian Splatting. Focusing exclusively on geometry reconstruction without color modeling, GeoGS also requires fewer primitives to represent the scene and is more robust to lighting variations. These advantages enable faster convergence in terms of geometric reconstruction.

\subsection{Depth and normal rendering}
\label{subsec:depth and normal rendering}
Based on the GeoGS primitives defined above, we first compute the depth and normal of an individual Gaussian primitive following \cite{huang20242d}.

For the $i$-th Gaussian primitive, let $\mathbf{t}_{x,i}$ and $\mathbf{t}_{y,i}$ denote the two orthogonal tangent directions, which are given by the first two columns of the rotation matrix. The corresponding normal direction $\mathbf{n}_i$ is therefore computed as:
\begin{equation}
\mathbf n_i= \mathbf{t}_{x,i} \times \mathbf{t}_{y,i}.
\label{eq:normal_define}
\end{equation}

The Gaussian disk is parameterized in the local tangent plane by its center point $\mathbf{\mu}_i\in\mathbb{R}^3$, two orthogonal tangential vectors $\mathbf{t}_{x,i}$ and $\mathbf{t}_{y,i}$, and scales $s_{x,i}$ and $s_{y,i}$. Let $\mathbf{R}_i=[\mathbf{t}_{x,i},\mathbf{t}_{y,i},\mathbf{n}_{i}]\in\mathbb{R}^{3\times3}$, $\mathbf{S}_i=\text{diag}(s_{x,i},s_{y,i},0)$, a point on the Gaussian disk with local coordinates $\mathbf x_i=(x_i,y_i)$ is then expressed in world coordinates as:
\begin{equation}
\mathbf{P}_i(\mathbf x)=\mathbf{\mu}_i+s_{x,i}\mathbf{t}_{x,i}x_i + s_{y,i}\mathbf{t}_{y,i}y_i=\mathbf{H}_i(x_i,y_i,1,1)^\top,
\label{eq:Puv}
\end{equation}

\begin{equation}
\mathbf{H}_i=\begin{bmatrix}s_{x,i}\mathbf{t}_{x,i} &s_{y,i}\mathbf{t}_{y,i}&0&\mathbf{\mu}_i \\ 0&0&0&1 \end{bmatrix}=\begin{bmatrix}\mathbf{R}_i\mathbf{S}_i&\mathbf{\mu}_i \\0&1 \end{bmatrix},
\label{eq:homogeneous}
\end{equation}
where $\mathbf{H}_i$ transforms a point from the local coordinate system of the Gaussian disk to the world coordinate system.

Let $\mathbf{W}\in\mathbb{R}^{4\times4}$ denote the projective transformation from world coordinates to screen coordinates. The image space points are obtained by:
\begin{equation}
\mathbf{X}=(uz_i,vz_i,z_i,z_i)^\top=\mathbf{W}\mathbf{P}_i(x_i,y_i)=\mathbf{W}\mathbf{H}_i(x_i,y_i,1,1)^\top,
\label{eq:screen_point}
\end{equation}
where $\mathbf{X}$ represents the ray corresponding to pixel $\mathbf{u}=(u, v)$ intersecting the 2D Gaussian splat at depth $z_i$. For each ray-splat intersection, the intersection coordinates $(x_i,y_i)$ in local tangent space and intersection depth $z_i$ are obtained by solving (\ref{eq:screen_point}) as detailed in \cite{huang20242d}.

The Gaussian value $\hat{\mathcal{G}}_i$ and contribution of the $i$-th Gaussian $\omega_i(\mathbf{u})$ at the intersection point are computed as:
\begin{equation}
\hat{\mathcal{G}}_i(\mathbf{u})=\exp(-\frac{x_i^2+y_i^2}{2}),
\label{eq:intersection_gaussian_value}
\end{equation}

\begin{equation}
\omega_i(\mathbf u)=\alpha_i\hat{\mathcal{G}}_i(\mathbf u)\prod_{j=1}^{i-1}(1-\alpha_j\hat{\mathcal{G}}_j(\mathbf u)),
\label{eq:2dgs_weight}
\end{equation}
where $\alpha_i$ and $\alpha_j$ denote the opacity of the $i$-th and $j$-th Gaussian disks.

We then render the depth and normal map as follows:
\begin{equation}
\hat{\mathbf{D}}(\mathbf{u}) =\frac{\sum\limits_{i} \omega_i(\mathbf{u}) z_i(\mathbf{u})}{\sum\limits_i \omega_i(\mathbf{u})} ,
\label{eq:depth_render}
\end{equation}

\begin{equation}
\hat{\mathbf{N}}(\mathbf{u}) =\sum_i \omega_i(\mathbf{u}) \mathbf{n}_i(\mathbf{u}),
\label{eq:normal_render}
\end{equation}
where $z_i(\mathbf{u})$ and $n_i(\mathbf{u})$ denote the depth and normal of the i-th Gaussian along the ray at pixel $\mathbf{u}$ as mentioned above.

\subsection{Loss function}
As GeoGS removes appearance-related parameters, conventional color-rendering loss (\ref{eq:color rendering loss}) is no longer available. We introduce a set of geometry-oriented loss functions to optimize the Gaussian primitives, consisting of single-view and multi-view objectives. The single-view losses directly supervise the rendered depth and normal. The multi-view losses further use the rendered depth to establish cross-view correspondences and derive geometry supervision from photometric consistency between neighboring RGB images. Together, these losses enable effective geometry optimization without appearance modeling.

\subsubsection{\textbf{Single-view losses}}
\label{subsubsec:single-view_loss}
The single-view losses optimize the GeoGS primitives using geometric consistency from the current rendering and the 3D geometric priors available from the SLAM frontend as supervision signals.

\textbf{Normal consistency.}
We adopt the normal consistency loss from PGSR\cite{chen2024pgsr} to encourage the Gaussian normals to agree with the surface normal estimated by the rendered depth map:
\begin{equation}
\mathcal{L}_{nc}=\frac{1}{|\Omega|}\sum_{\mathbf{u}\in\Omega}(1-|\nabla\mathbf{I}(\mathbf{u})|)(1-\hat{\mathbf{N}}(\mathbf{u})^\top\mathbf{N}_d(\mathbf{u})),
\label{normal_consistency}
\end{equation}
where $\Omega$ denotes the valid set of pixels, $\hat{\mathbf{N}}(\mathbf{u})$ denotes the rendered normal map at pixel $\mathbf{u}$, $\nabla\mathbf{I}(\mathbf{u})$ denotes the image gradient normalized to the range of 0 to 1. $\mathbf{N}_d(\mathbf{u})$ is computed by applying finite differences to the depth map. Specifically, let $\mathbf{P}_0$, $\mathbf{P}_1$, $\mathbf{P}_2$, and $\mathbf{P}_3$ be the back-projected 3D points corresponding to the left, right, top, bottom neighbors of pixel $\mathbf{u}$. Then:
\begin{equation}
\mathbf{N}_d(\mathbf{u})=\frac{(\mathbf{P}_1-\mathbf{P}_0)\times(\mathbf{P}_3-\mathbf{P}_2)}{\|(\mathbf{P}_1-\mathbf{P}_0)\times(\mathbf{P}_3-\mathbf{P}_2)\|_2}.
\label{eq:depth_normal}
\end{equation}

\textbf{Depth smoothness.}
We use an edge-aligned depth smoothness loss to encourage smooth surfaces while preserving sharp boundaries:
\begin{equation}
\mathcal{L}_{ds}=\frac{1}{|\Omega|}\sum_{\mathbf{u}\in\Omega}|\nabla\hat{\mathbf{D}}(\mathbf u)|\cdot e^{-\eta|\nabla\mathbf{I}(\mathbf u)|}+|\nabla\mathbf{I}(\mathbf u)|\cdot e^{-\eta|\nabla\hat{\mathbf{D}}(\mathbf u)|} ,
\label{eq:depth_smoothness}
\end{equation}
where $\nabla\mathbf{I}(\mathbf u)$ and $\nabla\hat{\mathbf{D}}(\mathbf{u})$ denote the gradients of RGB image and rendered depth at pixel $\mathbf u$, and $\eta$ is a weighting coefficient. The first term suppresses depth variations in regions with small image gradients, while the second term allows large depth variations around image boundaries, helping to preserve sharp geometric boundaries.

\textbf{Distortion loss.}
We also adopt the depth distortion loss in \cite{huang20242d}:
\begin{equation}
\mathcal{L}_{dist}=\frac{1}{|\Omega|}\sum_{\mathbf{u}\in\Omega}\sum_{i,j}\omega_i(\mathbf{u}) \omega_j(\mathbf{u})  |z_i(\mathbf{u})-z_j(\mathbf{u})|,
\label{eq:distortion}
\end{equation}
where $w_i(\mathbf{u})$ and $w_j(\mathbf{u})$ are the blending weights of the i-th and j-th intersections defined in (\ref{eq:2dgs_weight}), and $z_i(\mathbf{u})$ and $z_j(\mathbf{u})$ are the depth of the intersection points of the i-th and j-th Gaussian primitive at pixel $\mathbf{u}$ defined in Sec.~\ref{subsec:depth and normal rendering}.

\textbf{Depth and normal L1 losses.}
When dense depth and monocular normal estimates are available from the frontend (Sec.~\ref{subsec:frontend}), we further introduce direct L1 losses on depth and normal maps:
\begin{equation}
\mathcal{L}_d=\|\hat{\mathbf{D}}-\mathbf{D}_f\|_1,
\label{eq:depth_l1}
\end{equation}

\begin{equation}
\mathcal{L}_n=\|\hat{\mathbf{N}}-\mathbf{N}_f\|_1,
\label{eq:normal_l1}
\end{equation}
where $\hat{\mathbf{D}}$ and $\hat{\mathbf{N}}$ denote the rendered depth and normal, $\mathbf{D}_f$ and $\mathbf{N}_f$ denote the dense depth and monocular normal estimates from the frontend, respectively.

Finally, the single-view loss is defined as
\begin{equation}
\mathcal{L}_{single}=\lambda_{nc}\mathcal{L}_{nc}+\lambda_{ds}\mathcal{L}_{ds}+\lambda_{dist}\mathcal{L}_{dist}+\lambda_{d}\mathcal{L}_{d}+\lambda_{n}\mathcal{L}_{n},
\label{eq:single_loss}
\end{equation}
where $\lambda_{nc}$, $\lambda_{ds}$, $\lambda_{dist}$, $\lambda_{d}$ and $\lambda_{n}$ denote the contribution of each loss term.

\begin{figure}[t]
\centering
\includegraphics[width=0.49\textwidth]{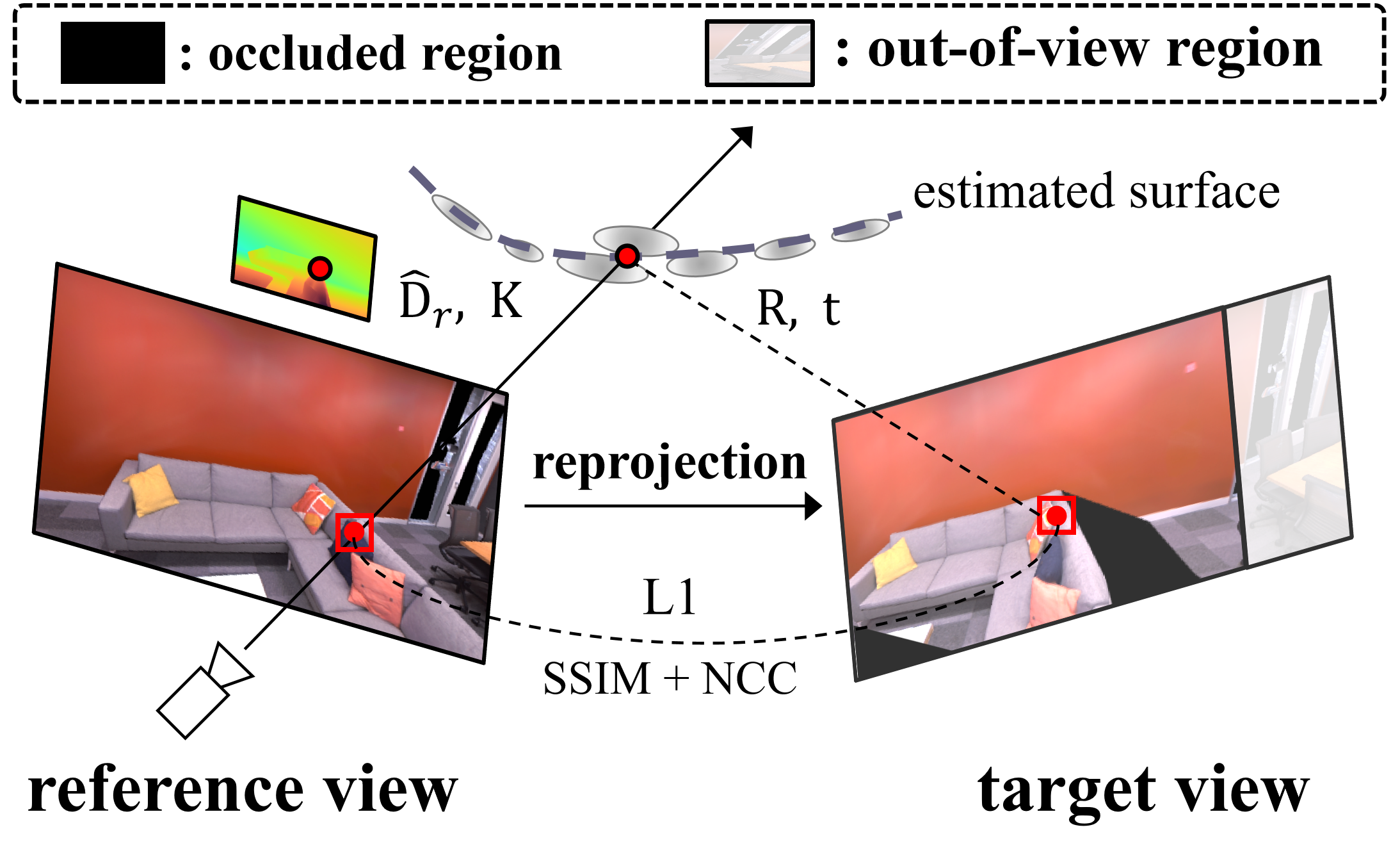}
\caption{Multi-view supervision in GeoGS mapping. The rendered depth map $\hat{\mathbf{D}}_r$ of is used to back-project pixels of reference view into 3D and reproject them to the target view. Visibility and image-boundary checks retain valid projections, over which L1, SSIM, and NCC losses provide cross-view photometric supervision for geometry optimization.}
\label{fig:multi-view}
\end{figure}

\begin{figure}[t]
\centering
\includegraphics[width=0.49\textwidth]{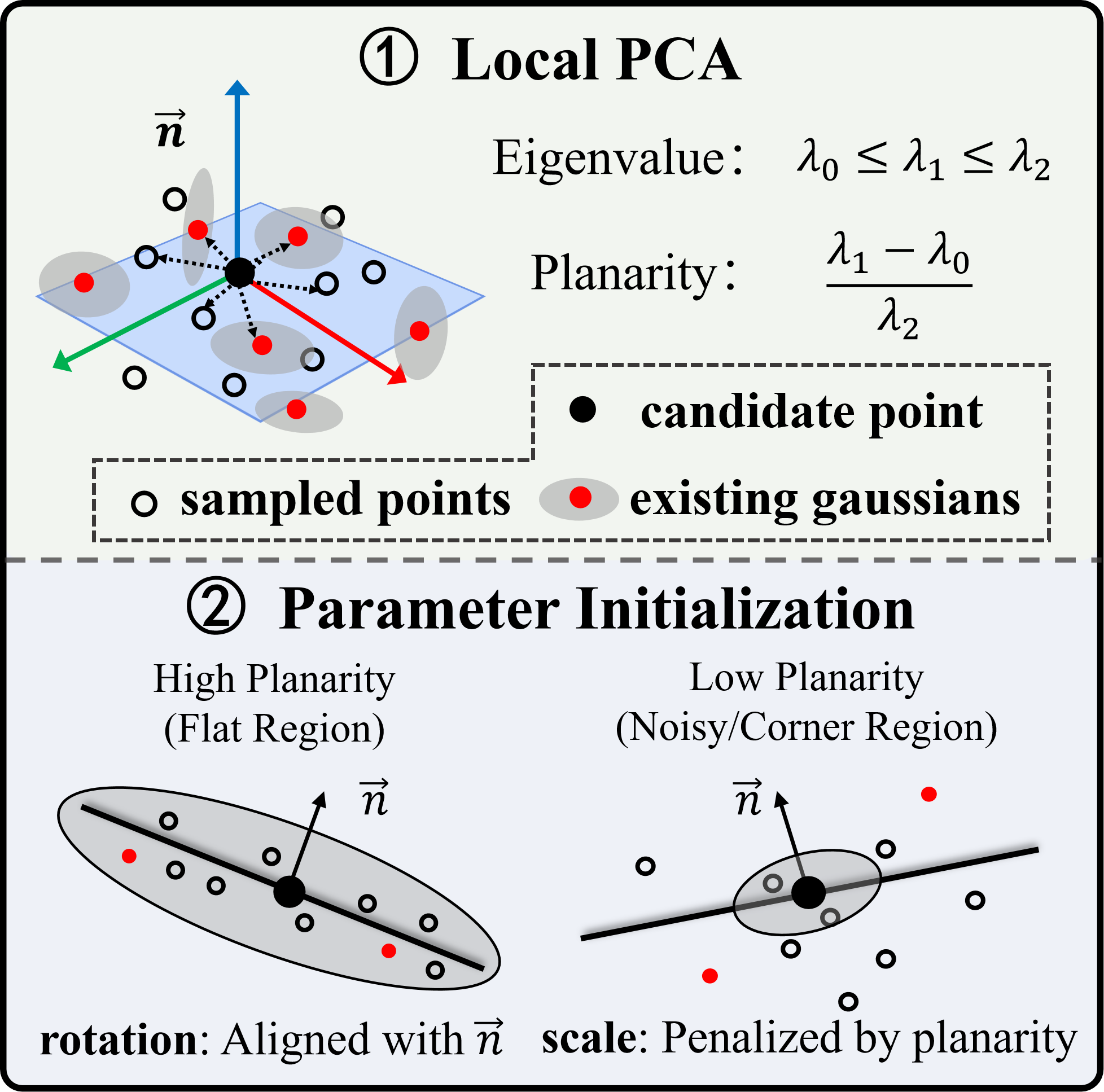}
\caption{Local-plane driven initialization for GeoGS primitives. For each candidate point, local PCA estimates the surface normal and planarity from neighboring points. The Gaussian rotation is initialized by aligning its normal with the estimated local-plane normal, while its scale is reduced in low-planarity neighborhoods to limit spatial support in geometrically uncertain regions.}
\label{fig:init_pca}
\end{figure}

\subsubsection{\textbf{Multi-view losses}}
\label{subsubsec:multi-view_loss}

As illustrated in Fig.~\ref{fig:multi-view}, let $\mathbf{I}_r$ and $\mathbf{I}_t$ denote the reference and target views, respectively. Given the rendered depth map $\hat{\mathbf{D}}_r$ of reference view, a pixel $\mathbf{p}_r$ in the reference view is projected onto the target view by
\begin{equation}
\mathbf{p}_t = \Pi\left( \mathbf{T}_{tr} \Pi^{-1} \left( \mathbf{p}_r, \hat{\mathbf{D}}_r(\mathbf{p}_r) \right) \right),
\label{eq:back_project}
\end{equation}
where $\mathbf{T}_{tr}$ denotes the transformation from the reference view to the target view, $\Pi(\cdot)^{-1}$ denotes back-projection from image coordinates to 3D space, and $\Pi(\cdot)$ denotes the perspective projection onto the image plane.

Since $\mathbf{p}_t$ generally lies at a sub-pixel location, the corresponding RGB value in the target view $\mathbf{I}_{r\rightarrow t}(\mathbf{p}_r)$ is obtained through bilinear interpolation:
\begin{equation}
\mathbf{I}_{r\rightarrow t}(\mathbf{p}_r)=
\mathbf{B}\left(\mathbf{I}_t,\mathbf{p}_t\right),
\label{eq:bilinear_interpolation}
\end{equation}
where $\mathbf{B}(\cdot)$ denotes the bilinear interpolation.

Due to occlusions and projections outside the target-image bounds, not all projected pixels provide valid correspondences. We then define an in-view mask $\mathbf{M}_{view}$ to exclude projections that fall outside the target image and an occlusion mask $\mathbf{M}_{occ}$ to filter out pixels that are not visible in the target view.

For the occlusion mask $\mathbf{M}_{occ}$, a projected pixel $\mathbf p_t$ is considered visible when its projected depth $\hat{\mathbf D}_{r\rightarrow t}(\mathbf p_t)$ is consistent with the rendered depth $\hat{\mathbf D}_t(\mathbf p_t)$ in the target view:
\begin{equation}
\mathbf M_{occ}=
\mathbf{1}\left(
\frac{
\left|
\hat{\mathbf D}_{r\rightarrow t}(\mathbf p_t)-
\hat{\mathbf D}_t(\mathbf p_t)
\right|
}{
\hat{\mathbf D}_{r\rightarrow t}(\mathbf p_t)
}
< \tau
\right),
\label{eq:occlusion_mask}
\end{equation}
where $\mathbf{1}(\cdot)$ denotes the indicator function, and $\tau$ is a predefined threshold.

The validation mask is then defined as
\begin{equation}
\mathbf{M}=\mathbf{M}_{view}\odot\mathbf{M}_{occ},
\label{eq:valid_mask}
\end{equation}
where $\odot$ denotes the element-wise multiplication.

Based on the reprojected image $\mathbf{I}_{r\rightarrow t}$, we compute L1, SSIM, and normalized cross-correlation (NCC) losses as follows:
\begin{equation}
\mathcal{L}_1=\left|\left| \mathbf{I}_r- \mathbf{I}_{r\rightarrow t} \right| \right| _1,
\label{eq:l1}
\end{equation}

\begin{equation}
\mathcal{L}_{ssim}=1-\mathrm{SSIM}\left( \mathbf{I}_r, \mathbf{I}_{r\rightarrow t} \right),
\label{eq:ssim}
\end{equation}

\begin{equation}
\mathcal{L}_{ncc}=1-\mathrm{NCC}\left( \mathbf{I}_r, \mathbf{I}_{r\rightarrow t} \right).
\label{eq:zncc}
\end{equation}

The multi-view loss is then defined as
\begin{equation}
\mathcal{L}_{mv}=
\lambda_1\mathcal{L}_1+\lambda_{ssim}\mathcal{L}_{ssim}+\lambda_{ncc}\mathcal{L}_{ncc},
\label{eq:multiview_loss}
\end{equation}
where $\lambda_1$, $\lambda_{ssim}$, and $\lambda_{ncc}$ denote the contribution of each loss, and the losses are computed over valid correspondences selected by the validation mask $\mathbf{M}$.

Finally, the loss function is defined as
\begin{equation}
\mathcal{L}_{geo}=\mathcal{L}_{single}+\mathcal{L}_{mv}.
\label{eq:geogs_loss}
\end{equation}

\subsection{Local-plane driven initialization}
As gaussian primitives gradually converge to the scene geometry during optimization, the initialization strategy can affect the efficiency of geometric convergence. To accelerate geometric convergence during incremental mapping, we introduce a local-plane driven initialization strategy for Gaussian primitives generated from each mapping frame.

Given the depth map of the current frame, we first back-project its valid pixels into 3D space and sample the resulting point cloud to obtain a set of candidate points. Each sampled point is subsequently used as the center of a newly initialized Gaussian primitive. Meanwhile, the existing Gaussian map contains the primitives initialized and optimized from previously processed frames. Their center positions therefore provide additional geometric context for estimating the local structure around the candidate points. 

As illustrated in Fig.~\ref{fig:init_pca}, for each candidate point, we construct a local neighborhood through a K-nearest neighbors (KNN) search over the union of the sampled points from the current frame and the centers of the existing Gaussians. During initialization from the first frame, the neighborhood is constructed solely from the sampled points back-projected from the depth map. We then estimate the local plane via Principal Component Analysis (PCA). Let $\lambda_0\leq\lambda_1\leq\lambda_2$ denote the eigenvalues of the local covariance matrix, and $\mathbf v_0$ denote the eigenvector corresponding to the smallest eigenvalue $\lambda_0$. The local planarity is defined as:

\begin{equation}
P=\frac{\lambda_1-\lambda_0}{\lambda_2}.
\label{eq:planarity}
\end{equation}

High planarity $P$ indicates that neighboring points are well approximated by a local plane, whereas low planarity suggests an ambiguous local planar structure. The estimated planarity is then used to refine Gaussian scale initialization.

The initial scale is first estimated from the average distance to neighboring points following 3DGS\cite{kerbl20233d}. For a candidate whose estimated planarity is below a predefined threshold, we further reduce its initialized scale by a factor of 10, thereby limiting the spatial support of Gaussians in geometrically uncertain regions. Meanwhile, the PCA-estimated normal $\mathbf v_0$ is used to initialize the Gaussian rotation by aligning the Gaussian normal with the local surface normal.

\begin{figure*}[t]
\centering
\includegraphics[width=1.0\textwidth]{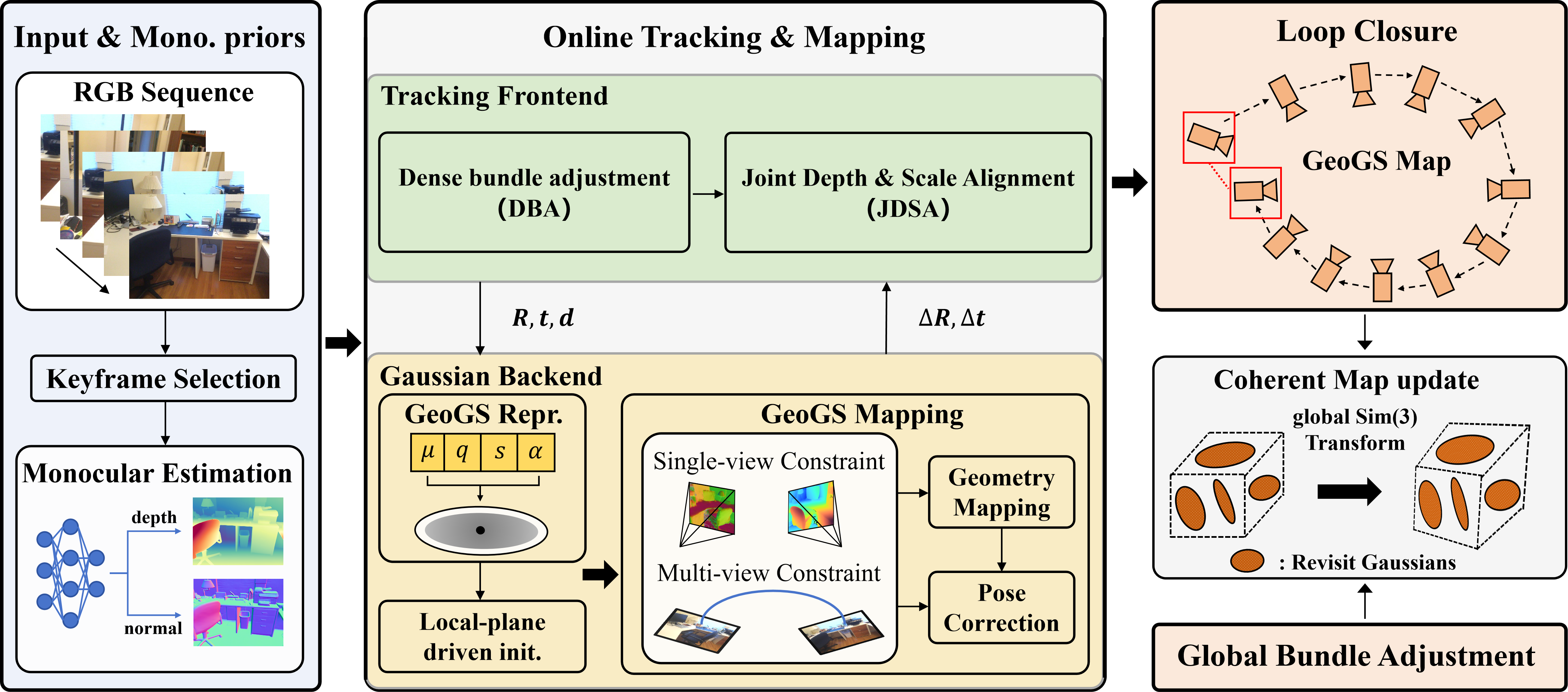}
\caption{Overview of the proposed GeoGS-SLAM system. Given a monocular RGB sequence, the frontend estimates camera poses and dense depth maps with monocular geometric priors. The Gaussian backend initializes GeoGS primitives using the proposed local-plane-driven strategy and optimizes the Gaussian map with single-view and multi-view supervision, followed by pose-only refinement over active keyframes. Global pose correction is performed through online loop closure with Sim(3)-based pose-graph bundle adjustment and a final global bundle adjustment. After each trajectory correction, the proposed coherent map update aligns the GeoGS map with the corrected poses while preserving local structural consistency.}
\label{fig:system_overview}
\end{figure*}

\section{GeoGS-SLAM}
In this section, we will introduce our GeoGS-SLAM system. As illustrated in Fig.~\ref{fig:system_overview}, the system mainly consists of frontend, Gaussian backend, and loop closure with coherent map update strategy. After the last keyframe is processed, the trajectory is further refined through a global bundle adjustment (BA) over all keyframes.

\subsection{Frontend}
\label{subsec:frontend}
The frontend of GeoGS-SLAM is built upon DROID-SLAM\cite{teed2021droid}, performing keyframe selection, camera tracking, and dense depth estimation.

An incoming frame is selected as a keyframe when the average flow distance to the latest keyframe, estimated by the optical flow network \cite{teed2020raft}, exceeds a predefined threshold. For each selected keyframe, we extract the monocular depth and normal priors through a pretrained network\cite{eftekhar2021omnidata}. The depth prior is then used to improve the depth estimation in the frontend, while the normal prior is used in the Gaussian backend to improve the mapping quality.

The system is initialized with the first 12 keyframes. The initialization performs dense bundle adjustment (DBA) on the keyframes. During online tracking, a frame graph $(\mathcal{V},\mathcal{E)}$ is maintained to represent the covisibility between frames. Each time a new keyframe is added, a local bundle adjustment (BA) is performed to estimate camera poses and depths. Specifically, given the predicted flow target $\mathbf{p}_{ij}^*$, the local BA can be formulated as 
\begin{equation}
\arg\min_{\mathbf{T},\mathbf{d}}\sum_{(i,j)\in \mathcal{E}}
\|\mathbf{p}_{ij}^*-\Pi(\mathbf{T}_{ij}\Pi^{-1}(\mathbf{p}_i,\mathbf{d}_i))\|
_{\Sigma_{ij}}^2, \Sigma_{ij}=\text{diag}\mathbf{w}_{ij},
\label{eq:DBA}
\end{equation}
where $(i,j)\in \mathcal{E}$ denotes that the corresponding images $\mathbf{I}_i$ and $\mathbf{I}_j$ share overlapping observations, $\mathbf{T}=\{\mathbf{T}_i\}$ and $\mathbf{d}=\{\mathbf{d}_i\}$ denote camera poses and depth maps, $\|\cdot\|_{\Sigma_{ij}}$ is the Mahalanobis distance weighting the error terms based on the confidence weights $\mathbf{w}_{ij}$ from the optical flow network, and $\mathbf{T}_{ij}$ denotes the relative pose.

To improve estimation in regions with inaccurate depth predictions, we further adopt the JDSA module\cite{zhang2025hi} that incorporates monocular depth priors into online tracking.

\subsection{Gaussian backend}
Given the camera poses and depth estimates provided by the frontend, the Gaussian backend optimizes the GeoGS map and performs local pose correction.

For each newly inserted keyframe, as introduced in Sec.~\ref{sec:geogs_mapping}, new GeoGS primitives are initialized from dense depth estimates in the frontend and integrated into the existing GeoGS map. The Gaussian backend then performs geometry optimization within a mapping sliding window consisting of recent keyframes together with two historical keyframes, accumulating losses over all window frames for back-propagation. The additional historical keyframes mitigate catastrophic forgetting during incremental mapping.

After the geometry optimization stage, the refined GeoGS map provides a geometrically consistent scene representation. We then perform a pose‑only correction that keeps all GeoGS primitives fixed and optimizes only the camera poses. Since the rendered depth, rendered normals, and multi-view reprojection process are all differentiable with respect to camera poses, they can be used as supervision for pose refinement:
\begin{equation}
\begin{aligned}
\min_{\mathbf T}\:\:\: & \mathcal{L}_{pose},\\
\mathcal{L}_{pose}=
\mathcal{L}_{d}^{pose}
+
&\mathcal{L}_{n}^{pose}
+
\mathcal{L}_{mv}^{pose},
\end{aligned}
\label{eq:pose_loss}
\end{equation}
where $\mathbf{T}$ denotes the camera poses in the frame graph. $\mathcal{L}_{d}^{pose}$, $\mathcal{L}_{n}^{pose}$, and $\mathcal{L}_{mv}^{pose}$ retain the same formulations as (\ref{eq:depth_l1}), (\ref{eq:normal_l1}), and (\ref{eq:multiview_loss}), respectively, while all GeoGS primitives remain fixed and only camera poses are optimized. 

Our refinement modifies only the camera poses and does not change the depth states maintained in the frontend frame graph. Since the frontend maintains a pose‑depth consistency using dense bundle adjustment (DBA), to ensure that the refined poses can be safely written back without breaking this consistency, we restrict pose correction exclusively to the active keyframes inside the current frame graph 
$(\mathcal{V},\mathcal{E)}$. To eliminate gauge freedom, we fix the pose of the earliest keyframe in the frame graph during this pose‑only refinement. The optimized poses are then written back to the frontend frame graph for subsequent tracking.

\subsection{Global pose correction}
To mitigate accumulated pose and scale drift, GeoGS-SLAM performs global pose correction through online loop closure and a final global bundle adjustment.

\textit{\textbf{1) Loop closure.}}

Loop closure detection is performed in parallel with online tracking to identify previously observed regions and introduce long-range constraints into the factor graph.

Following \cite{zhang2023hi}, for each new keyframe, candidate loop connections are searched among previously observed keyframes. A keyframe pair is accepted as a loop candidate when three conditions are satisfied: (i) the optical-flow distance between the two keyframes is below a predefined threshold, ensuring sufficient covisibility; (ii) the relative viewing orientation remains below a predefined threshold; (iii) their temporal separation exceeds a predefined threshold to exclude neighboring keyframes already constrained by local optimization. Once a valid loop candidate is detected, a loop edge is inserted into the factor graph. When loop candidates are identified, a Sim(3)-based pose-graph bundle adjustment (PGBA) is adopted to correct pose and scale drift.
\begin{figure*}[htbp]
    \centering
    \newlength{\imagewidth}
    \setlength{\labelwidth}{0.03\textwidth}
    \setlength{\imagewidth}{0.21\textwidth}
    \setlength{\tabcolsep}{0.3em}
    \renewcommand{\arraystretch}{1.2}
    
    \begin{tabular}{
        >{\centering\arraybackslash}m{\labelwidth}
        >{\centering\arraybackslash}m{\imagewidth}
        >{\centering\arraybackslash}m{\imagewidth}
        >{\centering\arraybackslash}m{\imagewidth}
        >{\centering\arraybackslash}m{\imagewidth}
    }
        & \textbf{Splat-SLAM} & \textbf{HI-SLAM2} & \textbf{GeoGS-SLAM} & \textbf{GT} \\
        
        \rotatebox[origin=c]{90}{\textbf{Room0}} &
        \includegraphics[width=\linewidth]{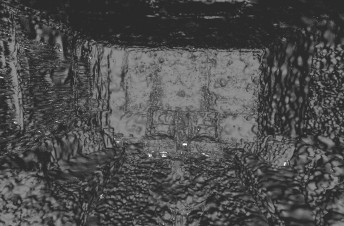} &
        \includegraphics[width=\linewidth]{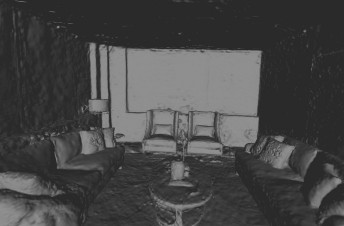} &
        \includegraphics[width=\linewidth]{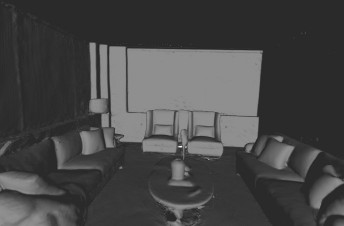} &
        \includegraphics[width=\linewidth]{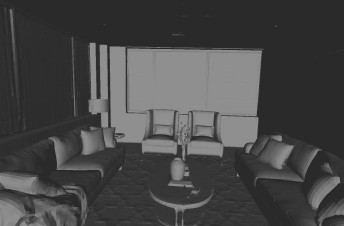} \\

        \rotatebox[origin=c]{90}{\textbf{Room2}} &
        \includegraphics[width=\linewidth]{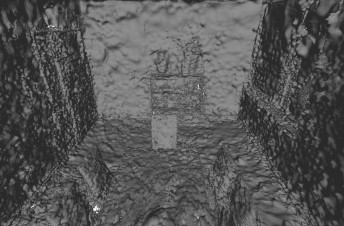} &
        \includegraphics[width=\linewidth]{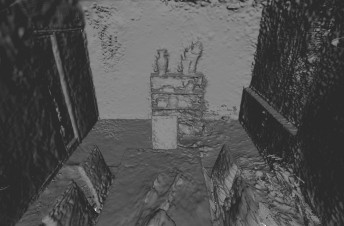} &
        \includegraphics[width=\linewidth]{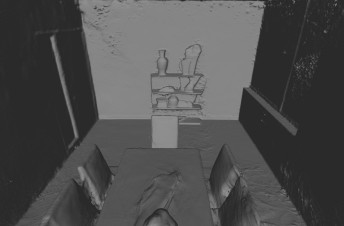} &
        \includegraphics[width=\linewidth]{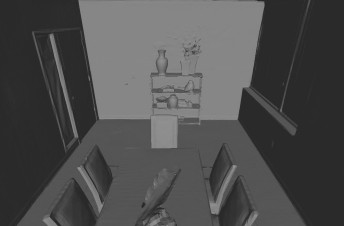} \\
        
        \rotatebox[origin=c]{90}{\textbf{Office0}} &
        \includegraphics[width=\linewidth]{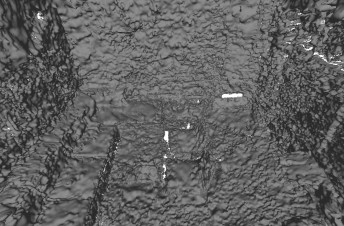} &
        \includegraphics[width=\linewidth]{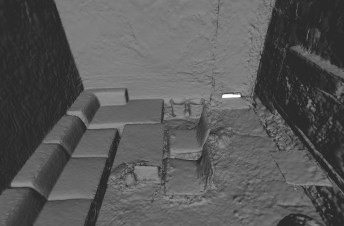} &
        \includegraphics[width=\linewidth]{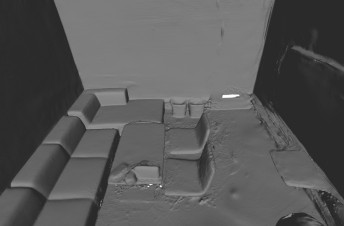} &
        \includegraphics[width=\linewidth]{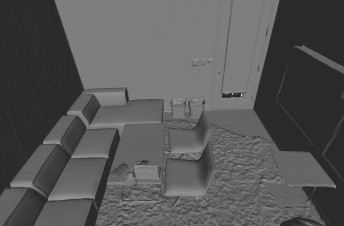} \\

        \rotatebox[origin=c]{90}{\textbf{Office2}} &
        \includegraphics[width=\linewidth]{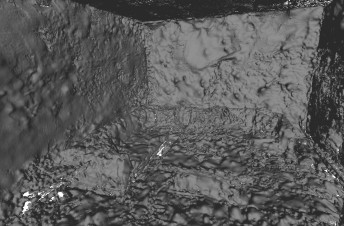} &
        \includegraphics[width=\linewidth]{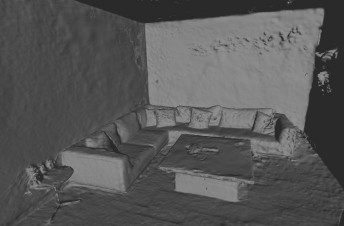} &
        \includegraphics[width=\linewidth]{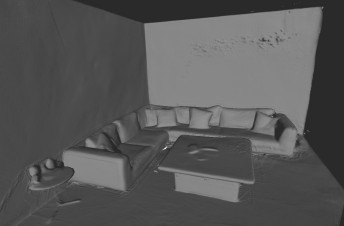} &
        \includegraphics[width=\linewidth]{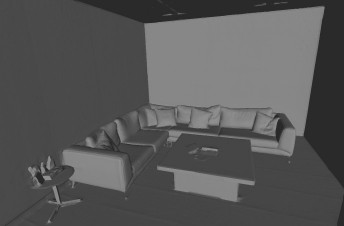} \\

        \rotatebox[origin=c]{90}{\textbf{Office4}} &
        \includegraphics[width=\linewidth]{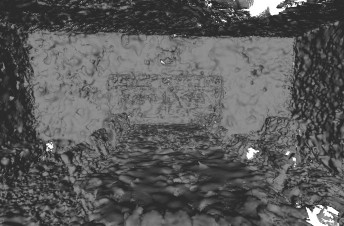} &
        \includegraphics[width=\linewidth]{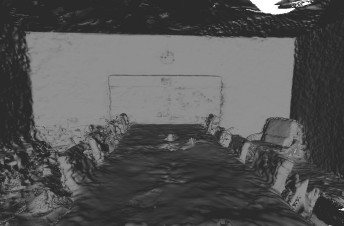} &
        \includegraphics[width=\linewidth]{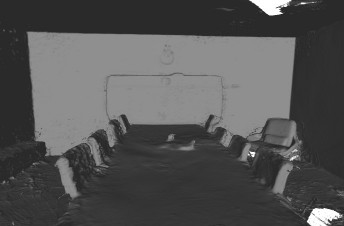} &
        \includegraphics[width=\linewidth]{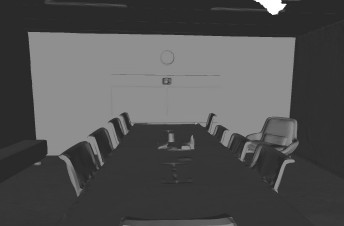} \\
    \end{tabular}
    
    \caption{Qualitative comparison of scene reconstruction on the Replica dataset.}
    \label{fig:replica_qualitative}
\end{figure*}

\textit{\textbf{2) Global bundle adjustment.}}

Although loop closure effectively mitigates accumulated drift through PGBA, the optimization is performed only over the sparse constraints maintained in the pose graph. Therefore, after all keyframes in a sequence have been processed, we perform a final global bundle adjustment to further refine the trajectory. Following DROID-SLAM\cite{teed2021droid}, dense bundle adjustment is performed on a global factor graph constructed from all keyframes.

\subsection{Coherent map update}
Both loop closure and global bundle adjustment modify the estimated camera trajectory, thereby introducing a discrepancy between the corrected poses and the existing GeoGS map. Because the Gaussian primitives are optimized under the original pose estimates, pose correction leaves the existing map inconsistent with the corrected poses, affecting global reconstruction consistency and subsequent incremental mapping. To avoid time-consuming retraining using revisited keyframes, previous works \cite{xu2024glc, zhang2025hi} associate each Gaussian with the keyframe from which the Gaussian was initialized, and propagate the pose correction to the associated Gaussians. Although efficient, they neglect spatial coherence: neighboring Gaussians that lie on the same surface but are associated with different keyframes may receive different, sometimes conflicting transformations, which may introduce tearing structure and floating artifacts as shown in Fig.~\ref{fig:ablation_map_update}.

To preserve structural consistency, we introduce a coherent map update strategy that estimates a unified Sim(3) transformation over the set of Gaussians associated with the revisited region. Specifically, instead of directly updating each Gaussian using the pose correction of its associated keyframe $i$, we first compute the target position $\mu_j'$ of each revisited Gaussian after pose correction:
\begin{equation}
\mathbf{\mu}_j'= (\mathbf T_i'^{-1} \mathbf T_i\mathbf{\mu_j})/scale_i,
\label{eq:target_position}
\end{equation}
where $scale_i$ denotes the scale correction during loop correction, $\mathbf T_i'$ and $\mathbf T_i$ denote the camera poses after and before the pose correction.

Then, a global Sim(3) transformation $s_g,\mathbf{R}_g,\mathbf{t}_g$ is computed by solving:
\begin{equation}
\arg\min_{s_g,\mathbf{R}_g,\mathbf{t}_g}
\sum_{j\in\mathcal{G}_{revis}}
\left\|
\mathbf{\mu}_j'
-
(s_g\cdot \mathbf{R}_g\mathbf{\mu}_j+\mathbf{t}_g)
\right\|_2^2 ,
\label{eq:global_sim3}
\end{equation}
where $\mathcal{G}_{revis}$ denotes the set of revisited Gaussians.

The revisited Gaussians are subsequently updated using the estimated global transformation as:
\begin{equation}
\begin{aligned}
\mathbf{\mu}_j'' &= s_g\cdot\mathbf{R}_g\cdot\mathbf{\mu}_j+\mathbf{t}_g,\\
R(\mathbf{q_j}'') &= \mathbf{R}_g\cdot R(\mathbf{q_j}),\\
s_j'' &= s_g\cdot s_j ,
\end{aligned}
\label{eq:Gaussian_update}
\end{equation}
where $R(\cdot)$ denotes the conversion from quaternion to rotation matrix, and $\mathbf{\mu}_j''$, $\mathbf{q}_j''$ and $s_j''$ denote the updated parameters.

All revisited Gaussians therefore receive a coherent correction, aligning the dense map with the trajectory after pose correction while better preserving local geometric consistency.

\begin{table*}[t]
\caption{Online reconstruction evaluation on the Replica dataset over eight sequences. Acc. and Comp. are reported in cm ($\downarrow$); Comp.Rat. is reported in \% ($\uparrow$). Methods marked with $*$ are evaluated in online mode without offline map refinement.}
\label{tab:replica_recon}
\centering
\begin{tabular}{|c|c|c|c|c|c|c|c|c|c|c|}
\hline
Method & Metric & Room0 & Room1 & Room2 & Office0 & Office1 & Office2 & Office3 & Office4 & Avg \\
\hline
\multirow{3}{*}{DROID-SLAM\cite{teed2021droid}} & Acc.$\downarrow$ & 3.51 & 2.80 & 3.67 & \textbf{2.57} & 2.32 & 3.78 & 3.80 & 4.19 & 3.33 \\
 & Comp.$\downarrow$ & 7.47 & 5.90 & 11.85 & 4.52 & 7.60 & 10.37 & 10.11 & 10.06 & 8.49 \\
 & Comp.Rat.$\uparrow$ & 71.05 & 76.31 & 65.75 & 78.71 & 67.72 & 66.64 & 64.24 & 67.74 & 69.77 \\
\hline
\multirow{3}{*}{NICER-SLAM\cite{zhu2024nicer}} & Acc.$\downarrow$ & 2.53 & 3.93 & 3.40 & 5.49 & 3.45 & 4.02 & 3.34 & 3.03 & 3.65 \\
 & Comp.$\downarrow$ & \textbf{3.04} & 4.10 & \textbf{3.42} & 6.09 & 4.42 & \textbf{4.29} & \textbf{4.03} & 3.87 & 4.16 \\
 & Comp.Rat.$\uparrow$ & \textbf{88.75} & 76.61 & \textbf{86.10} & 65.19 & 77.84 & 74.51 & \textbf{82.01} & \textbf{83.98} & 79.37 \\
\hline
\multirow{3}{*}{GO-SLAM\cite{zhang2023go}} & Acc.$\downarrow$ & 4.60 & 3.31 & 3.97 & 3.05 & 2.74 & 4.61 & 4.32 & 3.91 & 3.81 \\
 & Comp.$\downarrow$ & 5.56 & 3.48 & 6.90 & 3.31 & 3.46 & 5.16 & 5.40 & 5.01 & 4.79 \\
 & Comp.Rat.$\uparrow$ & 73.35 & 82.86 & 74.23 & 82.56 & \textbf{86.19} & 75.76 & 72.63 & 76.61 & 78.00 \\
\hline
\multirow{3}{*}{HI-SLAM\cite{zhang2023hi}} & Acc.$\downarrow$ & 3.21 & 3.74 & 3.16 & 3.87 & 2.60 & 4.62 & 4.25 & 3.53 & 3.62 \\
 & Comp.$\downarrow$ & 3.25 & \textbf{3.08} & 4.09 & 5.29 & 8.83 & 4.42 & 4.06 & \textbf{3.72} & 4.59 \\
 & Comp.Rat.$\uparrow$ & 86.99 & \textbf{87.19} & 80.82 & 72.55 & 72.44 & 80.90 & 81.04 & 82.88 & 80.60 \\
\hline
\multirow{3}{*}{Splat-SLAM*\cite{sandstrom2025splat}} & Acc.$\downarrow$ & 3.56 & 3.33 & 3.11 & 5.16 & 4.02 & 4.29 & 3.24 & 3.47 & 3.77 \\
 & Comp.$\downarrow$ & 4.57 & 4.41 & 4.07 & \textbf{3.00} & 4.76 & 4.74 & 4.44 & 5.52 & 4.43 \\
 & Comp.Rat.$\uparrow$ & 77.37 & 79.08 & 83.72 & 69.03 & 70.91 & 72.83 & 80.99 & 78.33 & 76.53 \\
\hline
\multirow{3}{*}{HI-SLAM2*\cite{zhang2025hi}} & Acc.$\downarrow$ & 2.00 & 2.03 & 5.09 & 2.75 & 1.78 & 2.74 & 2.53 & 3.60 & 2.82 \\
 & Comp.$\downarrow$ & 3.90 & 3.69 & 4.77 & 3.32 & 3.54 & 4.76 & 4.30 & 5.21 & 4.19 \\
 & Comp.Rat.$\uparrow$ & 85.26 & 85.05 & 80.84 & 82.84 & 84.11 & 79.59 & 80.92 & 79.92 & 82.31 \\
\hline
\multirow{3}{*}{\textbf{GeoGS-SLAM}} & Acc.$\downarrow$ & \textbf{1.63} & \textbf{1.47} & \textbf{2.10} & 2.70 & \textbf{1.63} & \textbf{1.94} & \textbf{2.36} & \textbf{2.89} & \textbf{2.09} \\
 & Comp.$\downarrow$ & 3.92 & 3.22 & 3.83 & 3.46 & \textbf{3.20} & 4.34 & 4.47 & 5.17 & \textbf{3.95} \\
 & Comp.Rat.$\uparrow$ & 86.77 & 86.44 & 85.09 & \textbf{83.05} & 85.97 & \textbf{81.92} & 80.08 & 80.02 & \textbf{83.67} \\
\hline
\end{tabular}
\end{table*}
\begin{table*}[t]
\caption{Online camera tracking and reconstruction results on the Scannet++ dataset over four sequences. ATE and CD are reported in cm($\downarrow$).  Method marked with $*$ is evaluated in online mode without offline map refinement.}
\label{tab:scannetpp_joint}
\centering
\begin{tabular}{|c|c|c|c|c|c|c|}
\hline
Method & Sequence & sequence1 & sequence2 & sequence3 & sequence4 & Avg \\
\hline
\multirow{2}{*}{HI-SLAM2*\cite{zhang2025hi}} & ATE$\downarrow$ & 3.96 & 3.06 & 9.20 & 3.03 & 4.81  \\
 & CD$\downarrow$ & 13.66 & 8.15 & 18.97 & 8.94 & 12.43 \\
\hline
\multirow{2}{*}{\textbf{GeoGS-SLAM}} & ATE$\downarrow$ & \textbf{3.93} & \textbf{3.06} & \textbf{7.64} & \textbf{3.02} & \textbf{4.41}  \\
 & CD$\downarrow$ & \textbf{8.47} & \textbf{7.04} & \textbf{12.51} & \textbf{6.67} & \textbf{8.67} \\
\hline
\end{tabular}
\end{table*}

\begin{table*}[t]
\caption{Online camera tracking results on the scannet dataset over eight sequences. ATE is reported in cm($\downarrow$). Methods marked with $*$ are evaluated in online mode without offline map refinement.}
\label{tab:scannet_tracking}
\centering
\begin{tabular}{|c|c|c|c|c|c|c|c|c|c|}
\hline
Method & 0000 & 0054 & 0059 & 0106 & 0169 & 0181 & 0207 & 0233 & Avg \\
\hline
NICE-SLAM\cite{zhu2022nice} & 12.00 & 20.90 & 14.00 & 7.90 & 10.90 & 13.40 & 6.20 & 9.00 & 11.8 \\
E-SLAM\cite{johari2023eslam} & 7.30 & 36.30 & 8.50 & 7.50 & 6.50 & 9.00 & \textbf{5.70} & \textbf{4.30} & 10.6 \\
GLORIE-SLAM\cite{zhang2024glorie} & 5.50 & 9.40 & 9.10 & \textbf{7.00} & 8.20 & 8.30 & 7.50 & 5.10 & 7.50 \\
Splat-SLAM\cite{sandstrom2025splat} & 5.57 & 9.50 & 9.11 & 7.09 & 8.26 & 8.39 & 7.53 & 5.17 & 7.58 \\
HI-SLAM2*\cite{zhang2025hi} & 5.20 & 8.50 & 8.27 & 7.58 & 8.01 & 6.87 & 6.90 & 5.15 & 7.06 \\
\textbf{GeoGS-SLAM} & \textbf{5.20} & \textbf{8.47} & \textbf{8.25} & 7.13 & \textbf{8.00} & \textbf{6.86} & 6.89 & 5.21 & \textbf{7.00} \\
\hline
\end{tabular}
\end{table*}

\begin{figure}[htbp]
    \centering
    \setlength{\labelwidth}{0.03\textwidth}
    \setlength{\imagewidth}{0.21\textwidth}
    \setlength{\tabcolsep}{0.3em}
    \renewcommand{\arraystretch}{1.2}
    
    \begin{tabular}{
        >{\centering\arraybackslash}m{\labelwidth}
        >{\centering\arraybackslash}m{\imagewidth}
        >{\centering\arraybackslash}m{\imagewidth}
    }
        & \textbf{HI-SLAM2} & \textbf{GeoGS-SLAM}  \\
        
        \rotatebox[origin=c]{90}{\textbf{  sequence1}} &
        \includegraphics[width=\linewidth]{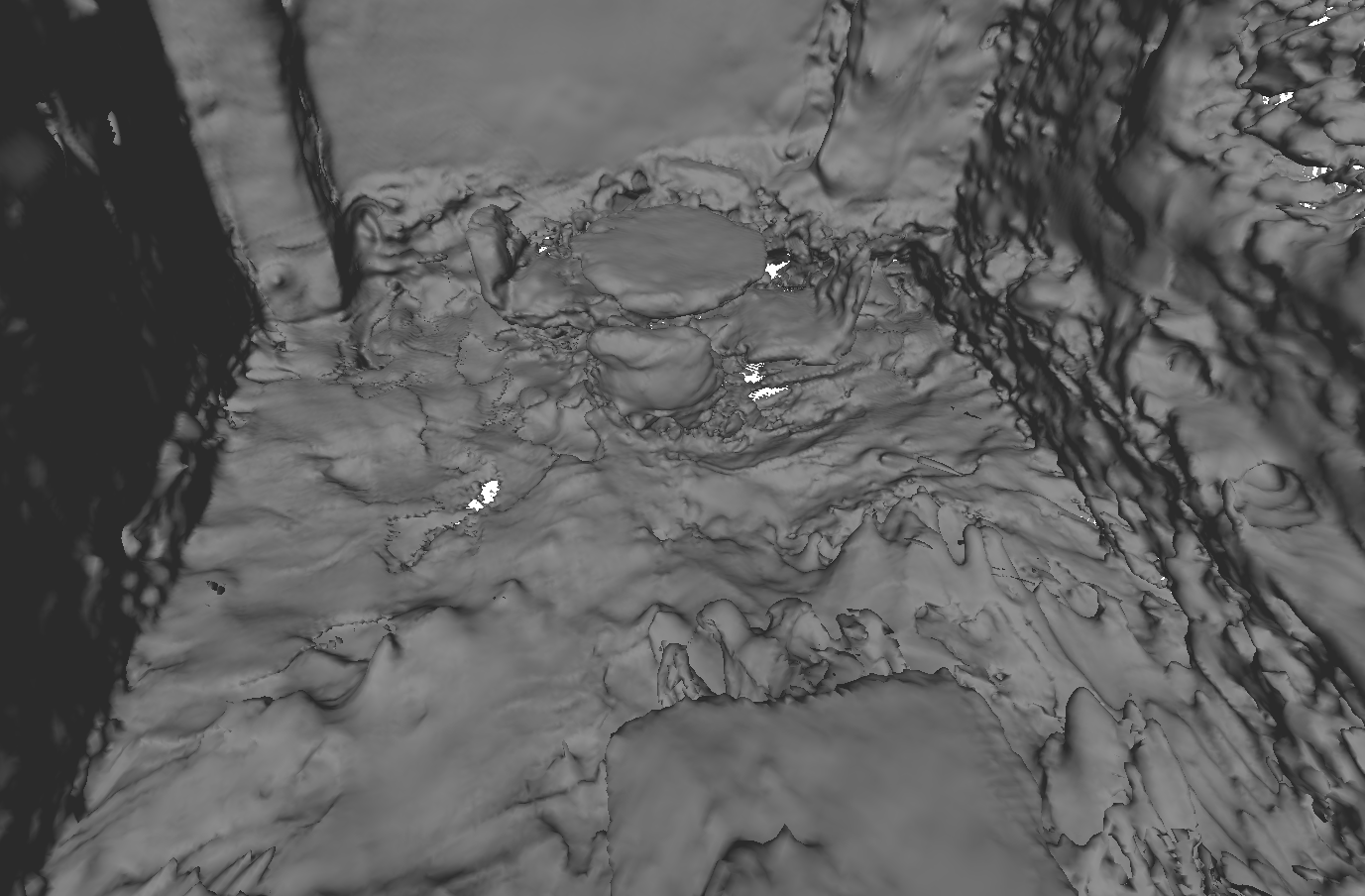} &
        \includegraphics[width=\linewidth]
        {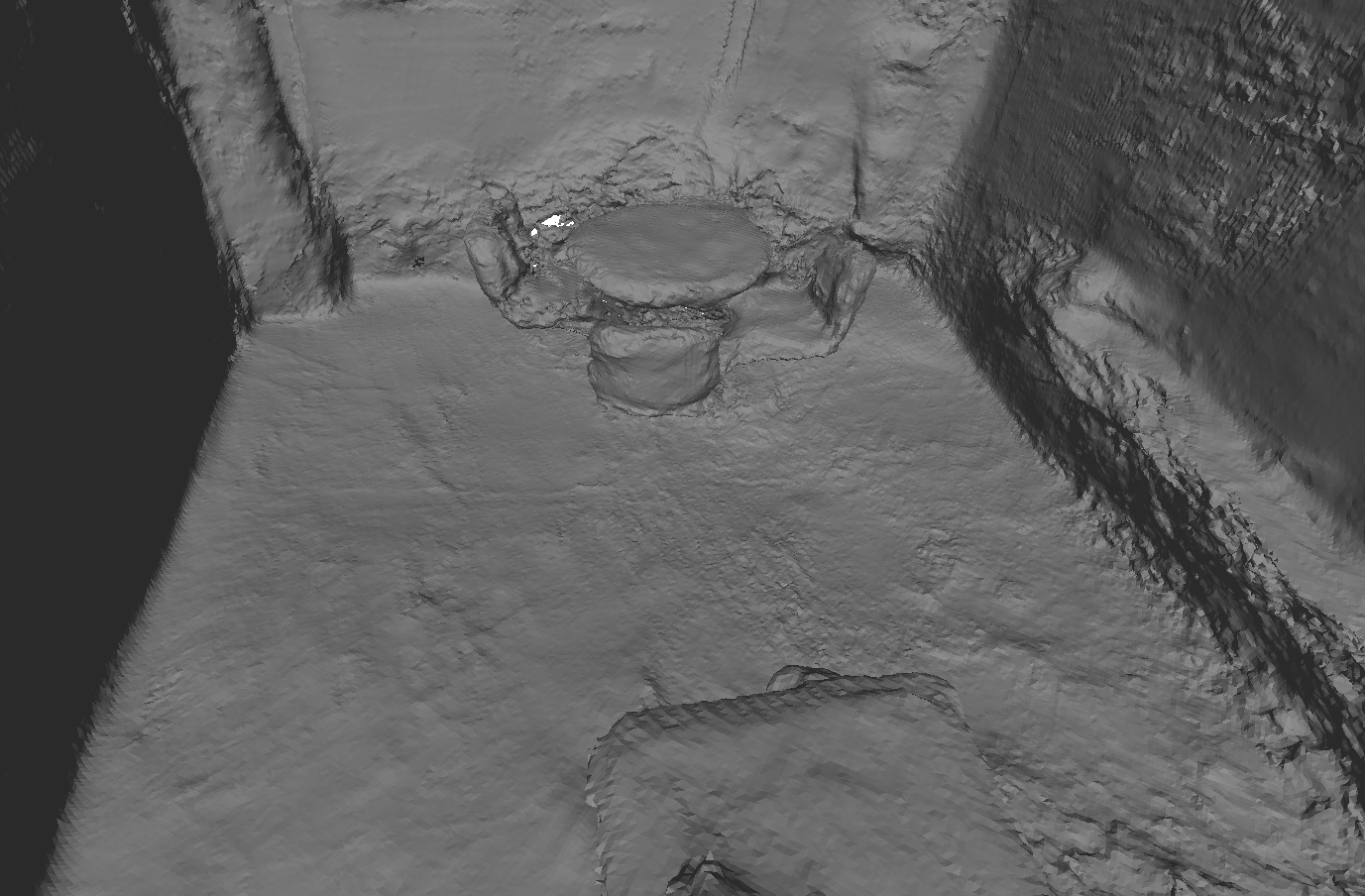} \\

        \rotatebox[origin=c]{90}{\textbf{  sequence4}} &
        \includegraphics[width=\linewidth]
        {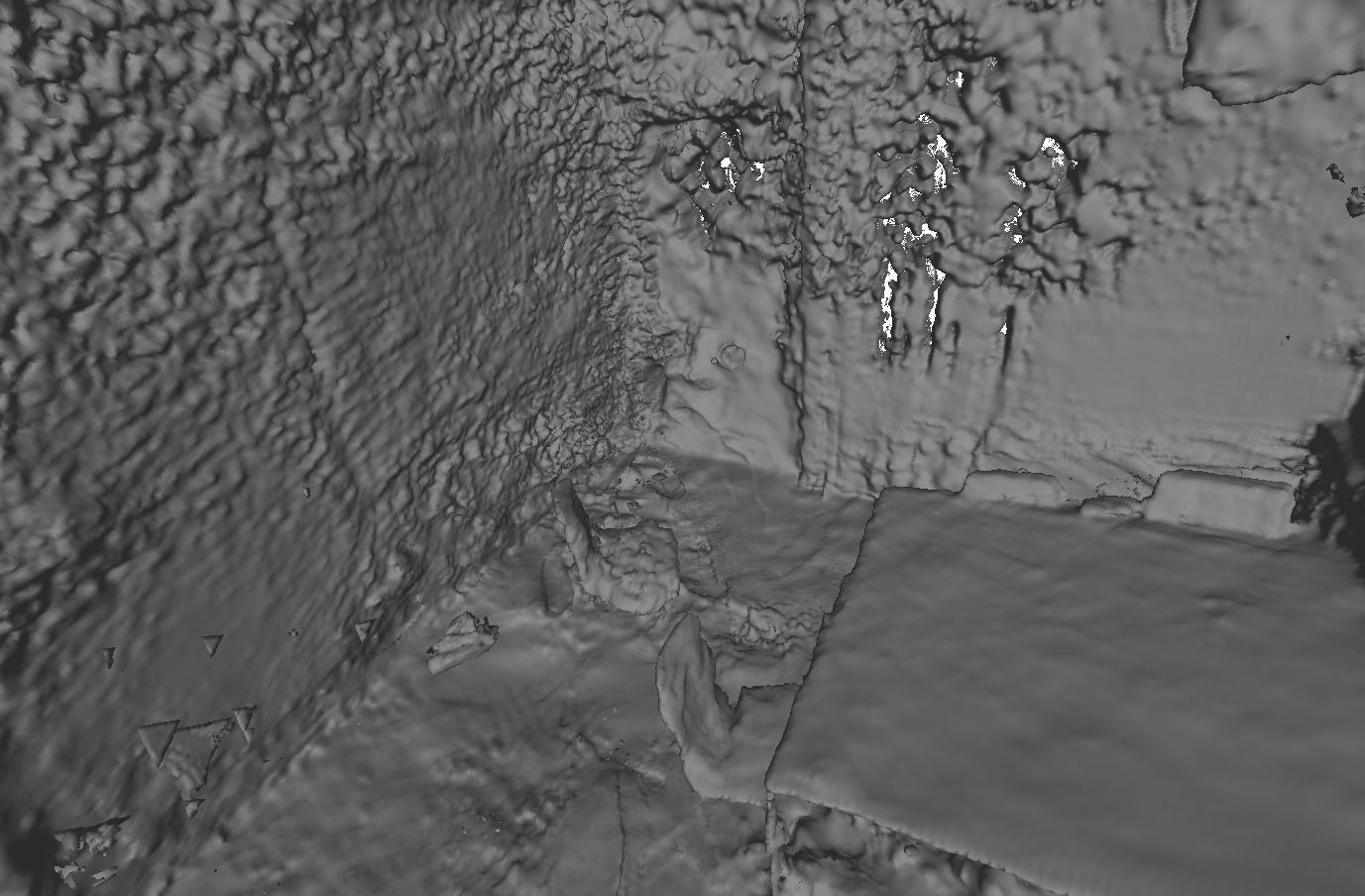} &
        \includegraphics[width=\linewidth]{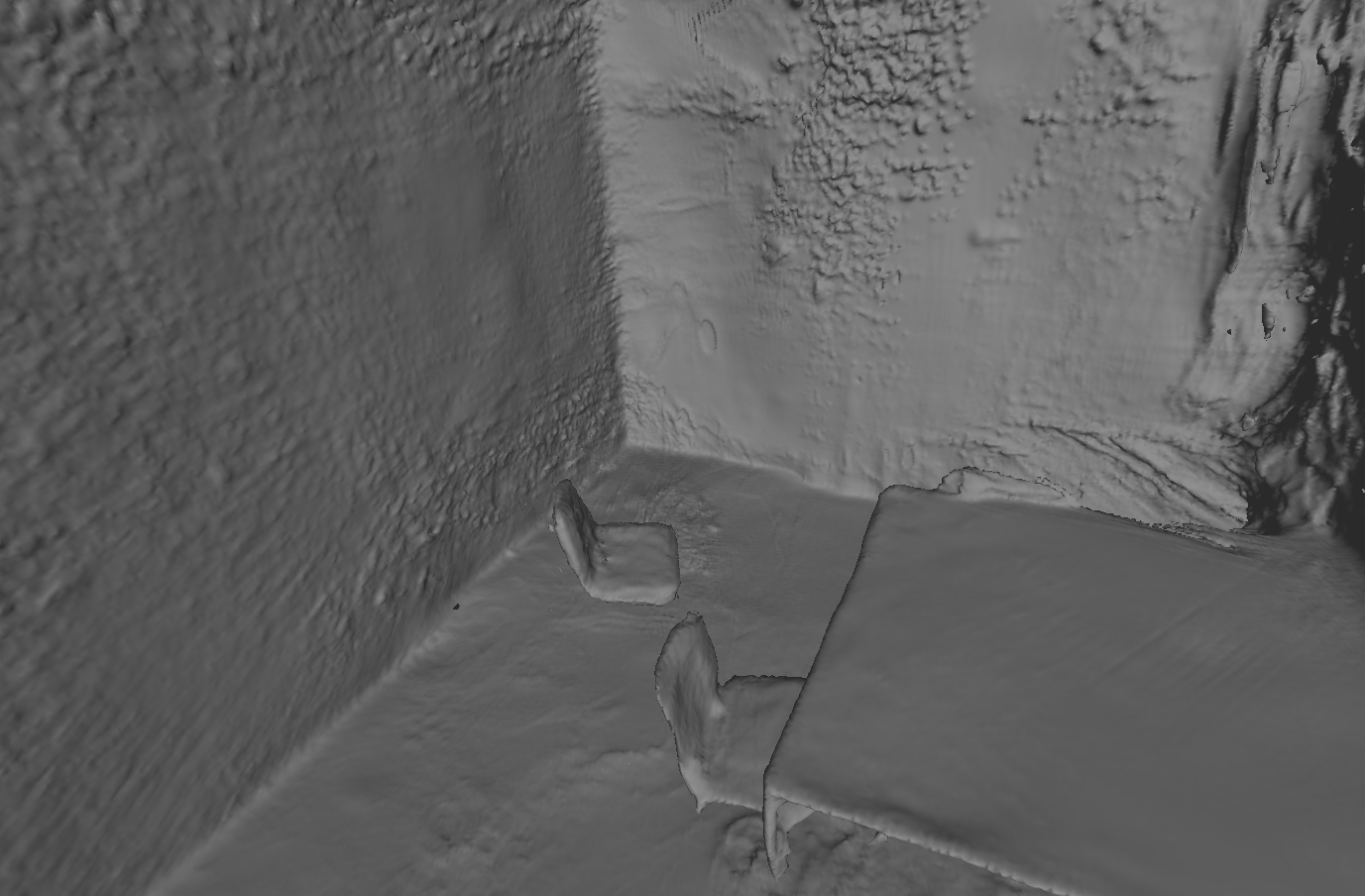} \\
        
    \end{tabular}
    
    \caption{Qualitative comparison of geometry reconstruction quality on the Scannet++ dataset. }
    \label{fig:scannet++_qualitative}
\end{figure}

\section{Experiments}
\subsection{Implementation Details}
\label{subsec:impl}
All experiments are performed on a workstation with an Intel Xeon Gold 6342 CPU and an NVIDIA RTX 4090 GPU with 24GB memory.  

Following Splat-SLAM\cite{sandstrom2025splat} and HI-SLAM2\cite{zhang2025hi}, the monocular depth and normal priors are extracted using Omnidata\cite{eftekhar2021omnidata}. For the evaluation of geometric reconstruction, we first obtain the meshes by fusing the rendered depth maps using TSDF Fusion algorithm\cite{werner2014truncated}, then we use the toolbox provided by Splat-SLAM\cite{sandstrom2025splat} to compute the metric. Additionally, we utilize Adam optimizer\cite{kingma2014adam} to optimize the parameters of Gaussian primitives.

\begin{figure}[htbp]
    \centering
    \setlength{\labelwidth}{0.03\textwidth}
    \setlength{\imagewidth}{0.21\textwidth}
    \setlength{\tabcolsep}{0.3em}
    \renewcommand{\arraystretch}{1.2}
    
    \begin{tabular}{
        >{\centering\arraybackslash}m{\labelwidth}
        >{\centering\arraybackslash}m{\imagewidth}
        >{\centering\arraybackslash}m{\imagewidth}
    }
        & \textbf{HI-SLAM2} & \textbf{GeoGS-SLAM}  \\
        
        \rotatebox[origin=c]{90}{\textbf{Scene0181}} &
        \includegraphics[width=\linewidth]{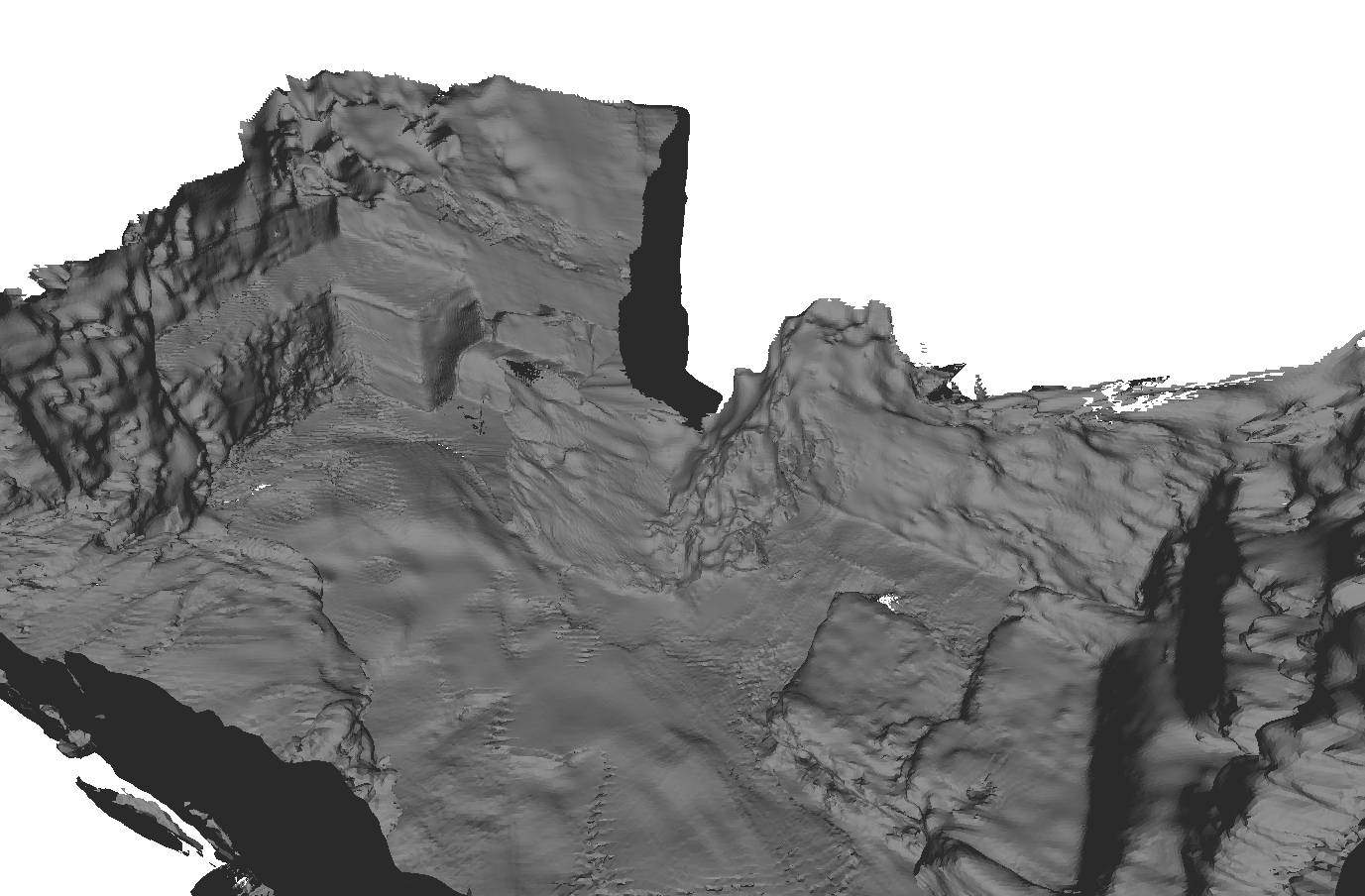} &
        \includegraphics[width=\linewidth]
        {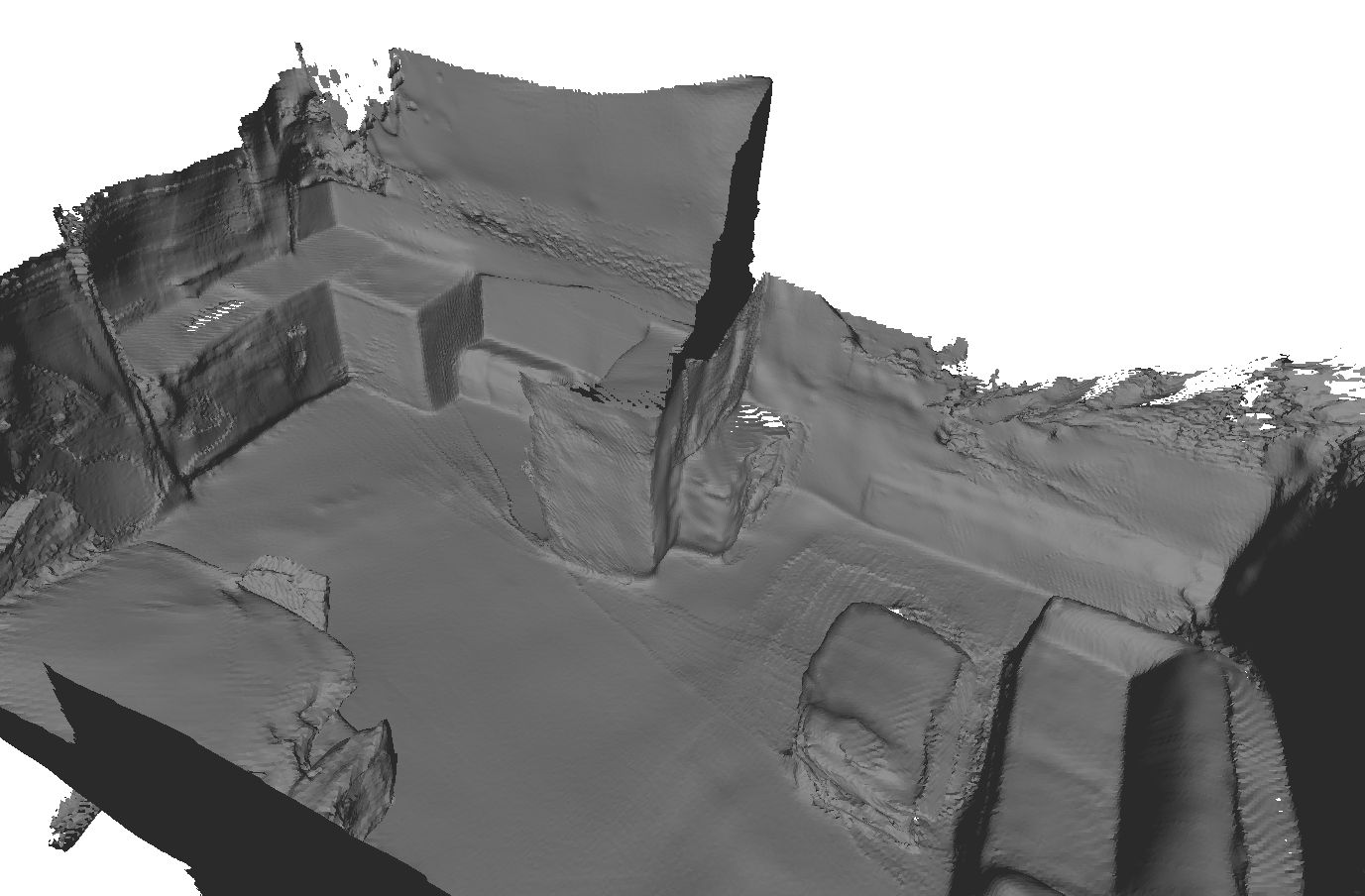} \\

        \rotatebox[origin=c]{90}{\textbf{Scene0233}} &
        \includegraphics[width=\linewidth]
        {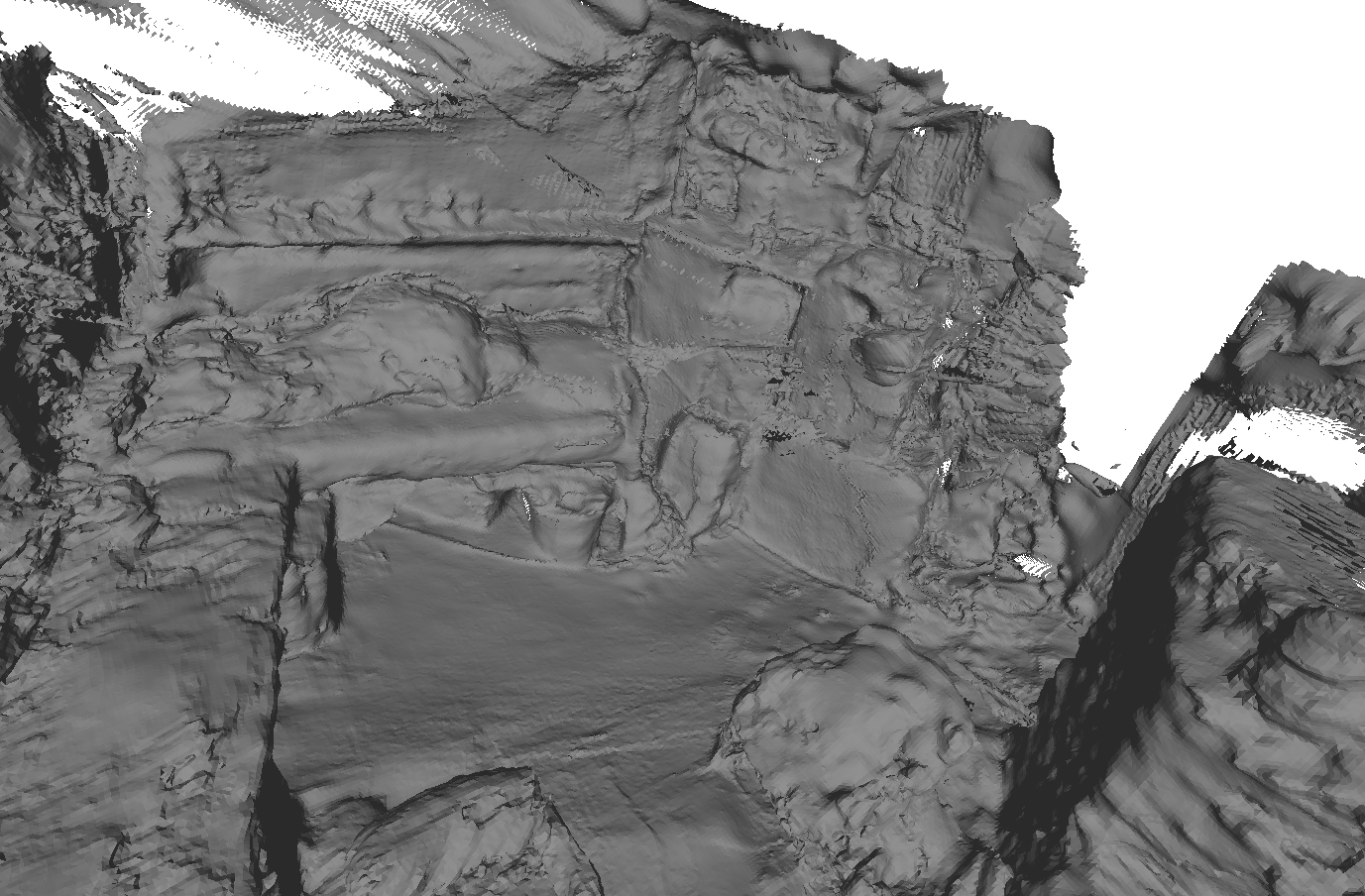} &
        \includegraphics[width=\linewidth]{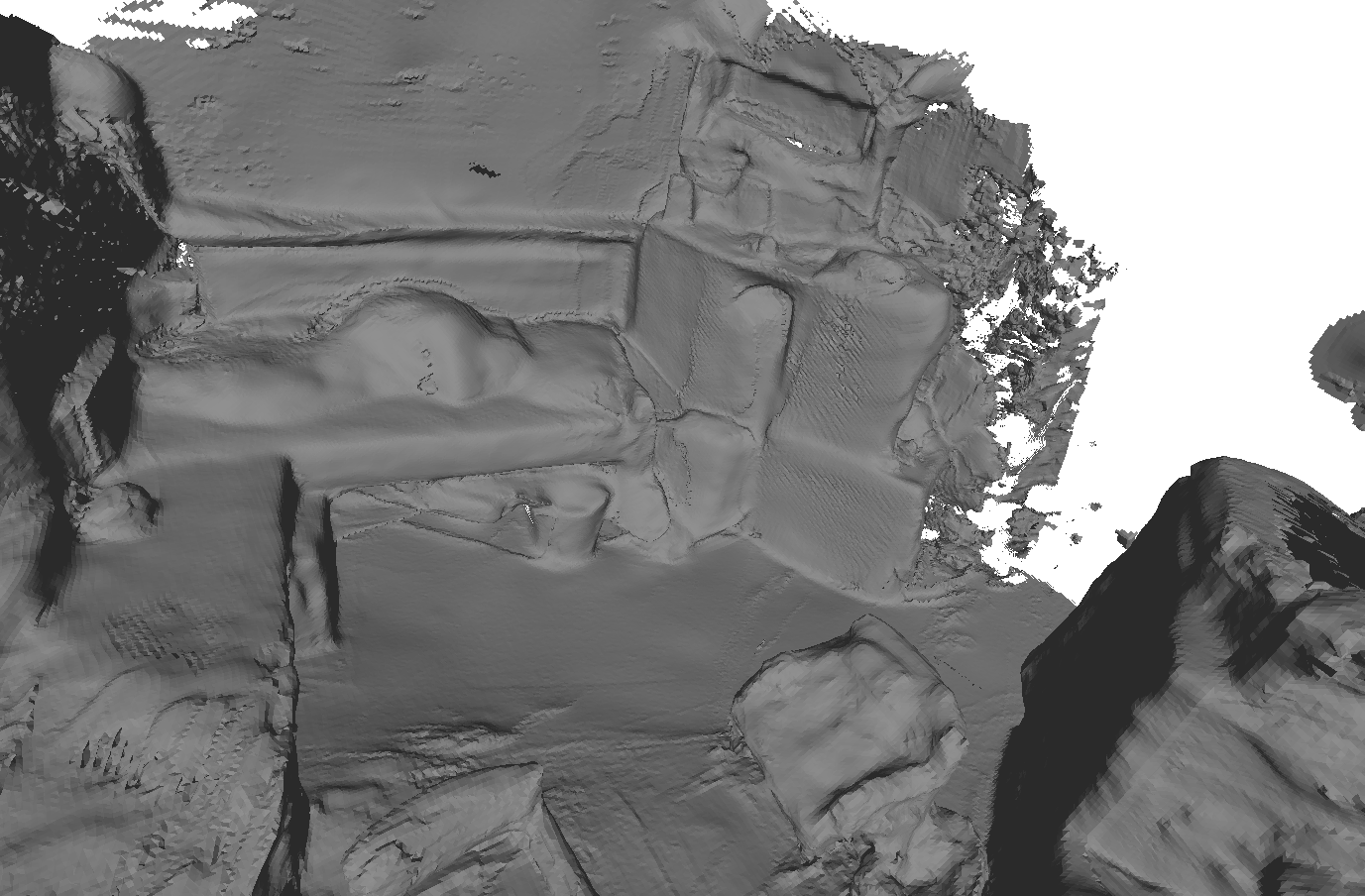} \\
        
    \end{tabular}
    
    \caption{Qualitative comparison of scene reconstruction on the Scannet dataset.}
    \label{fig:scannet_qualitative}
\end{figure}

\begin{table*}[t]
\caption{Per-scene chamfer distance on 15 DTU scans under a 2-minute optimization budget (CD, cm$\downarrow$).}
\label{tab:dtu_main}
\centering
\begin{tabular}{|c|c|c|c|c|c|c|c|c|c|c|c|c|c|c|c|c|}
\hline
Method & 24 & 37 & 40 & 55 & 63 & 65 & 69 & 83 & 97 & 105 & 106 & 110 & 114 & 118 & 122 & Avg \\
\hline
2DGS & 1.29 & 1.18 & 1.22 & 0.83 & 2.43 & 2.03 & 1.19 & 1.54 & 1.53 & 1.20 & 1.09 & 2.05 & 1.07 & 1.04 & 1.04 & 1.38 \\

PGSR & 2.22 & 1.78 & 2.56 & 2.72 & 3.66 & 2.77 & 1.88 & \textbf{1.09} & 1.80 & 2.42 & 0.75 & 3.08 & 2.89 & 0.97 & 0.49 & 2.07 \\

QGS & 2.15 & 1.34 & 1.65 & 1.45 & 2.75 & 2.33 & 1.49 & 1.32 & 1.94 & 1.20 & 1.85 & 2.27 & 1.81 & 1.26 & 1.18 & 1.73 \\

\textbf{GeoGS} &\textbf{ 0.53} & \textbf{0.89} & \textbf{0.36} & \textbf{0.34} & \textbf{0.96} & \textbf{0.66} & \textbf{0.56} & 1.25 & \textbf{0.96} & \textbf{0.62} & \textbf{0.46} & \textbf{0.62} & \textbf{0.34} & \textbf{0.42} & \textbf{0.41} & \textbf{0.62} \\
\hline
\end{tabular}
\end{table*}
\begin{table}[t]
\caption{Online camera tracking results on the Replica dataset. We report the mean ATE (cm$\downarrow$) over eight sequences in the online stage. Methods marked with $*$ are evaluated in online mode without offline map refinement. }
\label{tab:replica_tracking}
\centering
\begin{tabular}{|c|c|c|c|}
\hline
Method & ATE.$\downarrow$ \\
\hline
DROID-SLAM\cite{teed2021droid} & 0.38 \\
MGS-SLAM\cite{zhu2024mgs} & 0.32 \\
Splat-SLAM*\cite{sandstrom2025splat} & 0.38 \\
HI-SLAM2*\cite{zhang2025hi} & 0.32 \\
GSO-SLAM\cite{yeon2026gso} & 0.46 \\
\textbf{GeoGS-SLAM} & \textbf{0.28} \\
\hline
\end{tabular}
\end{table}

\subsection{Datasets}
\label{subsec:datasets}
\textbf{Replica.}
Replica\cite{straub2019replica} is a synthetic indoor benchmark with photorealistic $1200{\times}680$ RGB images, accurate camera trajectories, and dense mesh ground truth. We use eight sequences (Room0--Room2 and Office0--Office4) from \cite{sucar2021imap} for system-level SLAM evaluation, assessing both tracking accuracy and geometry reconstruction quality.

\textbf{ScanNet++.}
ScanNet++\cite{yeshwanth2023scannet++} is a real-world indoor benchmark with accurate geometric ground truth and diverse scene layouts. We evaluate online tracking and mapping performance using four representative sequences, 8b5caf3398 (sequence1), 39f36da05b (sequence2), 6d89a7320d (sequence3), fb564c935d (sequence4), from the iPhone captures subset. In addition, ground-truth meshes are available for the evaluation of geometric reconstruction. 

\textbf{ScanNet.}
ScanNet\cite{dai2017scannet} is a real-world indoor benchmark widely used for trajectory evaluation. Following\cite{sandstrom2025splat,zhang2025hi}, we choose a subset of 8 sequences to evaluate the tracking performance of GeoGS-SLAM.

\textbf{DTU.}
DTU\cite{jensen2014large} is a widely used multi-view stereo benchmark. Following 2DGS\cite{huang20242d}, we evaluate on a subset of 15 sequences to validate the effectiveness of the GeoGS representation and the proposed training framework under a time-constrained optimization setting, as detailed in Sec.~\ref{subsec:rep_val}.

\begin{figure*}[t]
\centering
\includegraphics[width=0.98\textwidth]{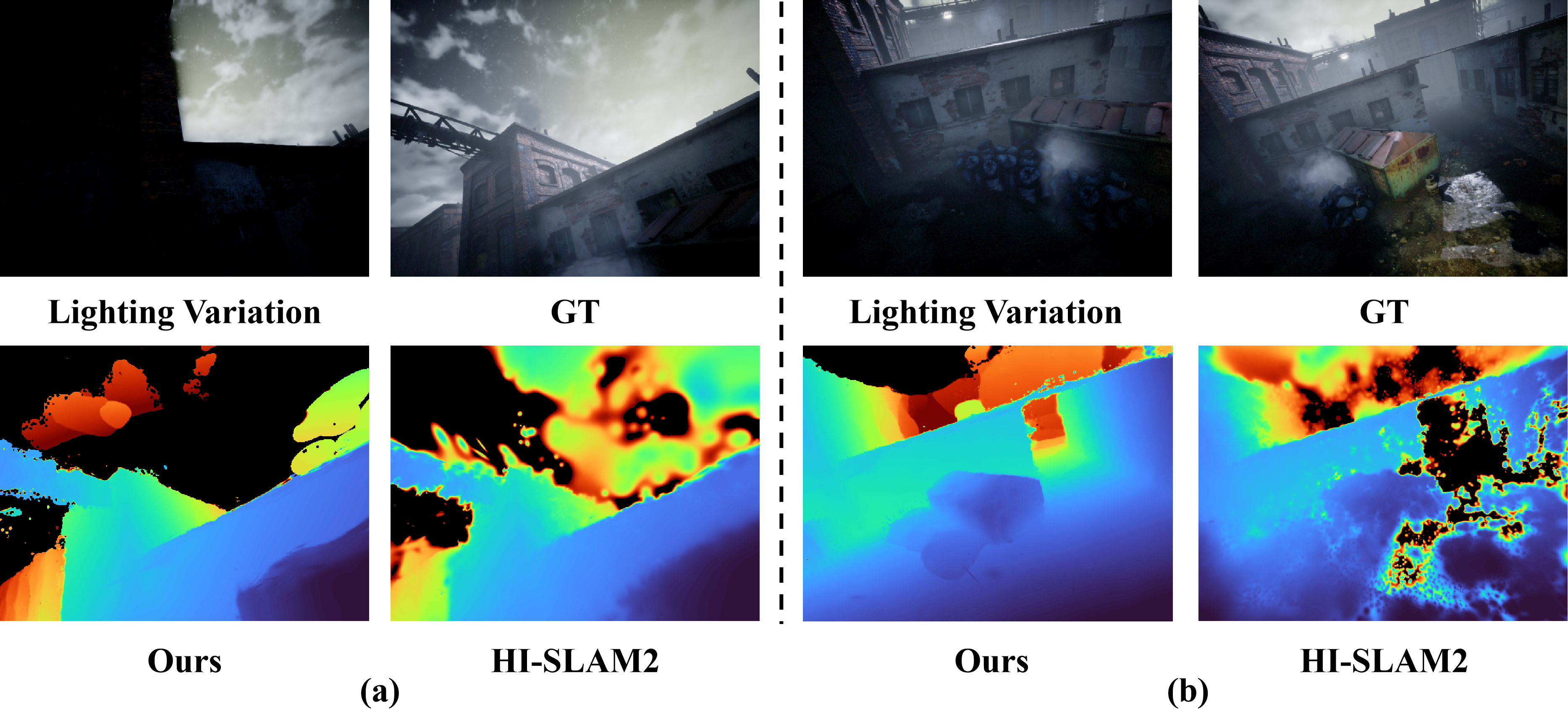}
\caption{Robustness of GeoGS-SLAM to illumination changes. (a) and (b) show two representative scene regions under different lighting conditions, with RGB observations in the top row and the corresponding reconstructed depth maps in the bottom row. Despite substantial appearance variations, GeoGS-SLAM preserves consistent scene geometry.}
\label{fig:illumination}
\end{figure*}

\begin{figure}[t]
\centering  
\includegraphics[width=0.49\textwidth]{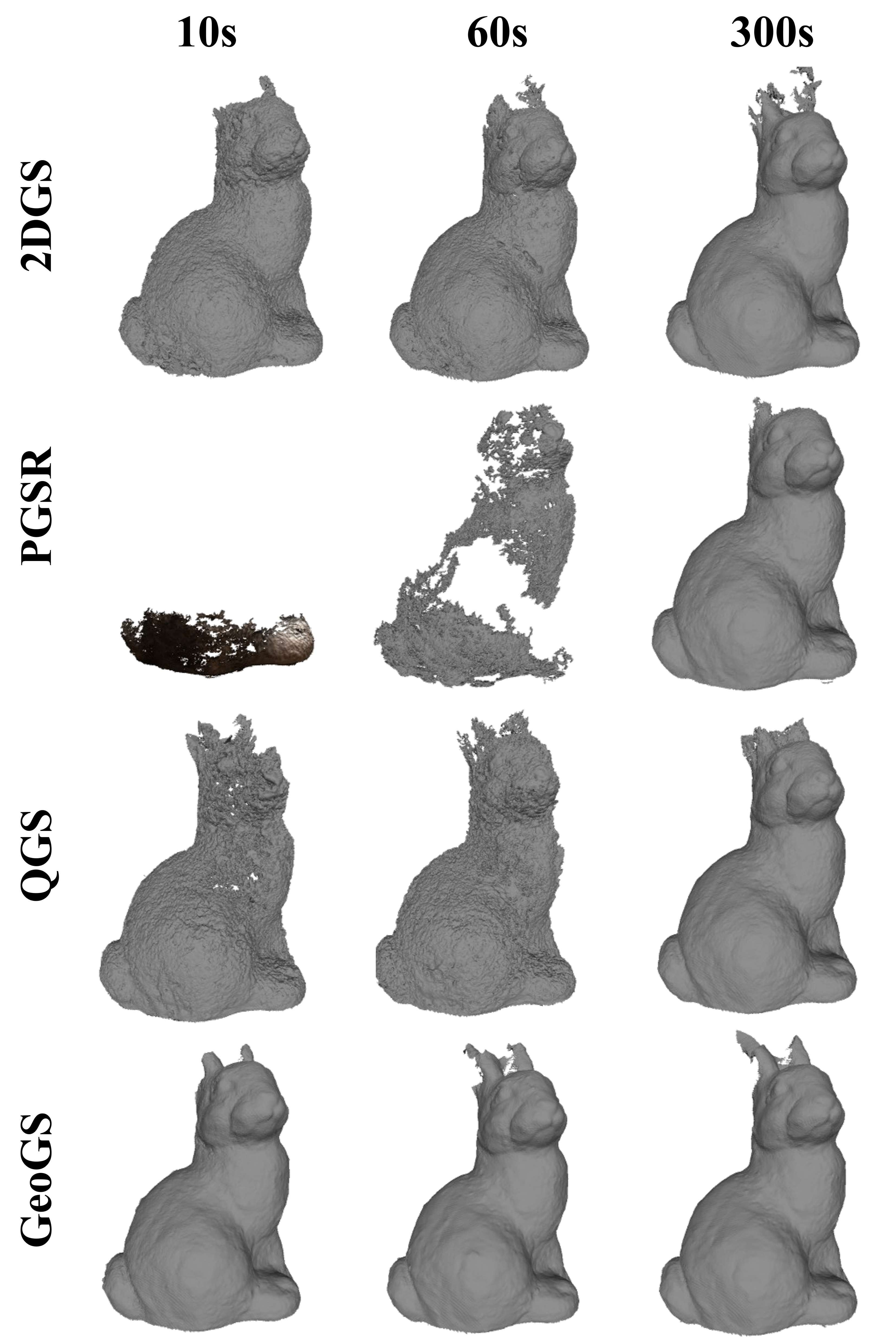}
\caption{Geometry convergence performance on the DTU Scan55 sequence. GeoGS achieves convergence in geometry reconstruction much faster than appearance-coupled methods.}
\label{fig:dtu55-convergence}
\end{figure}

\begin{figure}[t]
\centering
\includegraphics[width=0.49\textwidth]{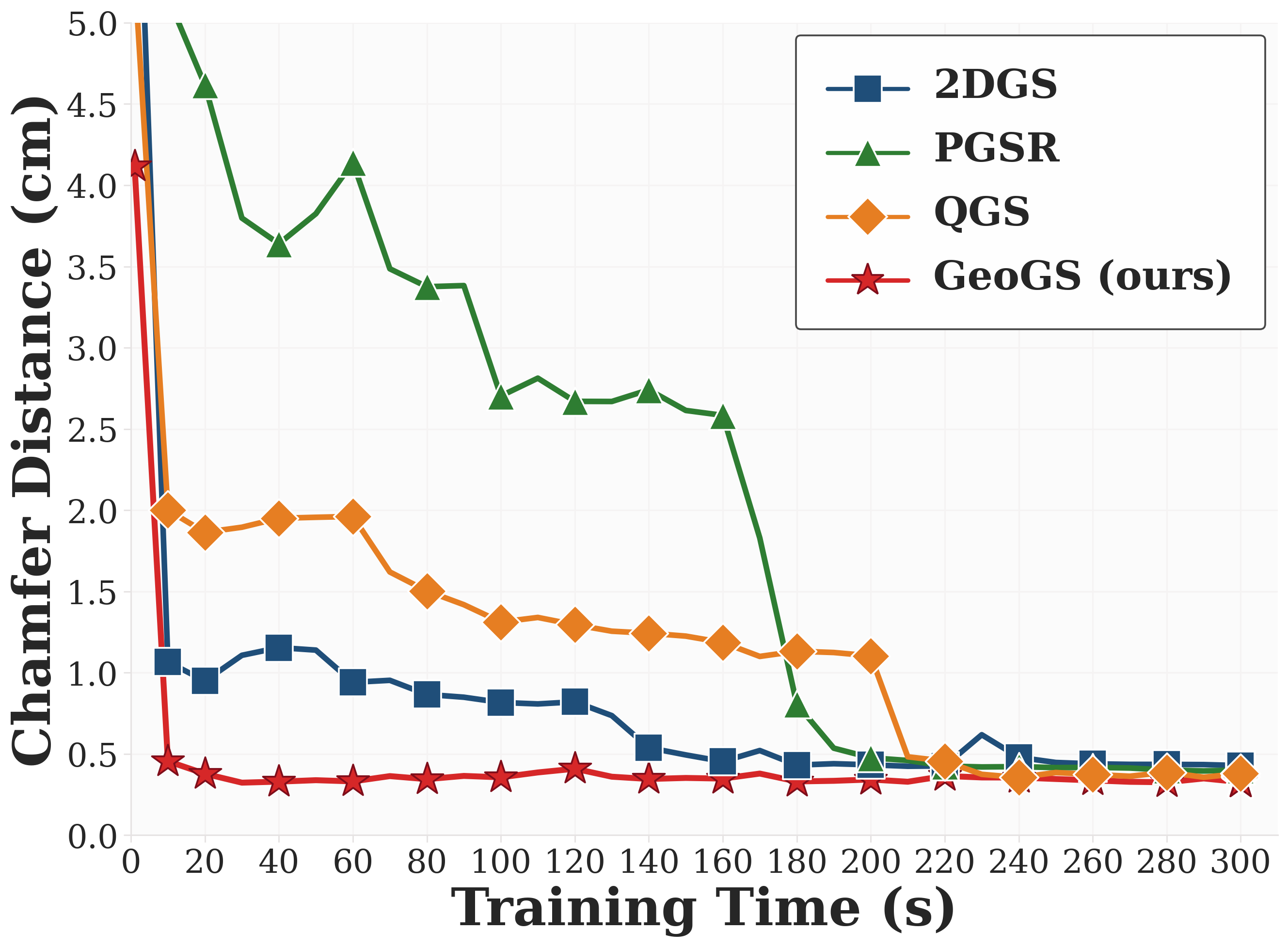}
\caption{Geometry convergence curves on the DTU Scan55 sequence measured by Chamfer Distance (CD).}
\label{fig:plot-convergence-dtu}
\end{figure}

\begin{figure}[t]
\centering
\includegraphics[width=0.49\textwidth]{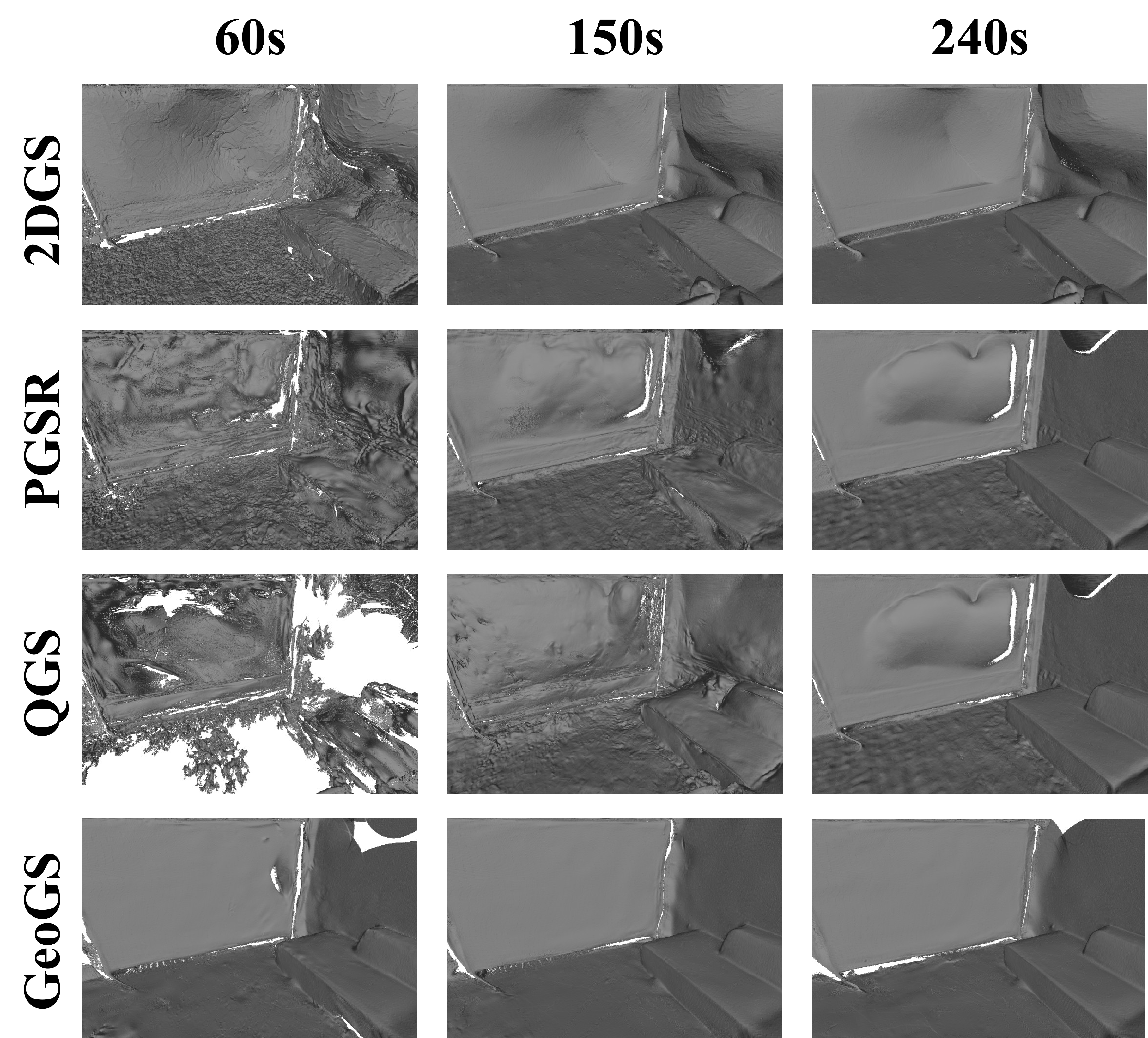}
\caption{Geometry convergence performance on the Replica Office0 sequence.}
\label{fig:replica-office0-convergence}
\end{figure}

\begin{table}[t]
\caption{Model efficiency comparison on the DTU Scan55 sequence under the time-constrained setting with 2-minute optimization budget.}
\label{tab:param_eff}
\centering
\begin{tabular}{|c|c|c|c|}
\hline
Method & Gaussians $\downarrow$ & Storage $\downarrow$ & CD $\downarrow$ \\
\hline
2DGS & 161\,k & 37.5\,MB & 0.83\,cm \\
PGSR & 231\,k & 54.6\,MB & 2.72\,cm \\
QGS & 200\,k & 8.1\,MB & 1.45\,cm \\
\textbf{GeoGS } & \textbf{23\,k} & \textbf{1.2} & \textbf{0.34} \\
\hline
\end{tabular}
\end{table}

\subsection{Metrics and Baselines}
\label{subsec:metrics_baselines}
For the evaluation of GeoGS-SLAM, we compare our method against representative SOTA methods, including DROID-SLAM\cite{teed2021droid}, NeRF-based SLAM systems NICER-SLAM\cite{zhu2024nicer}, GO-SLAM\cite{zhang2023go}, E-SLAM\cite{johari2023eslam}, GLORIE-SLAM\cite{zhang2024glorie}, HI-SLAM\cite{zhang2023hi} and GS-based SLAM systems MGS-SLAM\cite{zhu2024mgs}, Splat-SLAM\cite{sandstrom2025splat}, GSO-SLAM\cite{yeon2026gso} and HI-SLAM2\cite{zhang2025hi}.

On the Replica dataset, reconstruction quality is evaluated using mean Accuracy (Acc., cm$\downarrow$), mean Completeness (Comp., cm$\downarrow$), and Completeness Ratio (Comp.Rat., \%$\uparrow$, threshold = 5\,cm), together with tracking performance. On the ScanNet++ dataset, we report Chamfer Distance (CD, cm$\downarrow$) and tracking performance. On the ScanNet dataset, only quantitative tracking performance is evaluated. 

Specifically, let $\mathcal{P}$ and $\mathcal{G}$ denote the point sets sampled from the reconstructed and ground-truth point clouds, respectively. The metrics are computed as follows:
\begin{equation}
\mathrm{Acc}.=\frac{1}{|\mathcal P|}
\sum_{\mathbf{p}\in\mathcal P}
\min_{\mathbf{g}\in\mathcal G}
\|\mathbf{p}-\mathbf{g}\|^2 ,
\label{eq:acc}
\end{equation}

\begin{equation}
\mathrm{Comp.}=\frac{1}{|\mathcal G|}
\sum_{\mathbf{g}\in\mathcal G}
\min_{\mathbf{p}\in\mathcal P}
\|\mathbf{p}-\mathbf{g}\|^2 ,
\label{eq:comp}
\end{equation}

\begin{equation}
\mathrm{Comp.Rat.}=
\frac{1}{|\mathcal G|}
\sum_{\mathbf{g}\in\mathcal G}
\mathbf{1}\left(
\min_{\mathbf{p}\in\mathcal P}
\|\mathbf{p}-\mathbf{g}\|^2<\tau
\right)
\times100\%,
\label{eq:comp.rat}
\end{equation}
where $\mathbf{p}\in\mathcal{P}$ and $\mathbf{g}\in\mathcal{G}$ denote points in the reconstructed and ground-truth point sets, respectively, $\tau$ denotes the distance threshold, and $\mathbf{1(\cdot )}$ denotes the indicator function. Combining $\mathrm{Acc.}$ and $\mathrm{Comp.}$, Chamfer Distance (CD) is defined as:

\begin{equation}
\mathrm{CD}=\frac{1}{2}(\mathrm{Acc.+Comp.})
\label{eq:cd}
\end{equation}

Camera tracking accuracy is measured using the Absolute Trajectory Error (ATE, cm$\downarrow$), computed with EVO\cite{grupp2017evo}.

For the evaluation of GeoGS reconstruction, we compare our method against 2DGS\cite{huang20242d}, PGSR\cite{chen2024pgsr}, and QGS\cite{zhang2025quadratic} under the same time-constrained evaluation setting. Chamfer Distance (CD, cm$\downarrow$) is used as the evaluation metric.

\begin{figure}[t]
\centering
\includegraphics[width=0.49\textwidth]{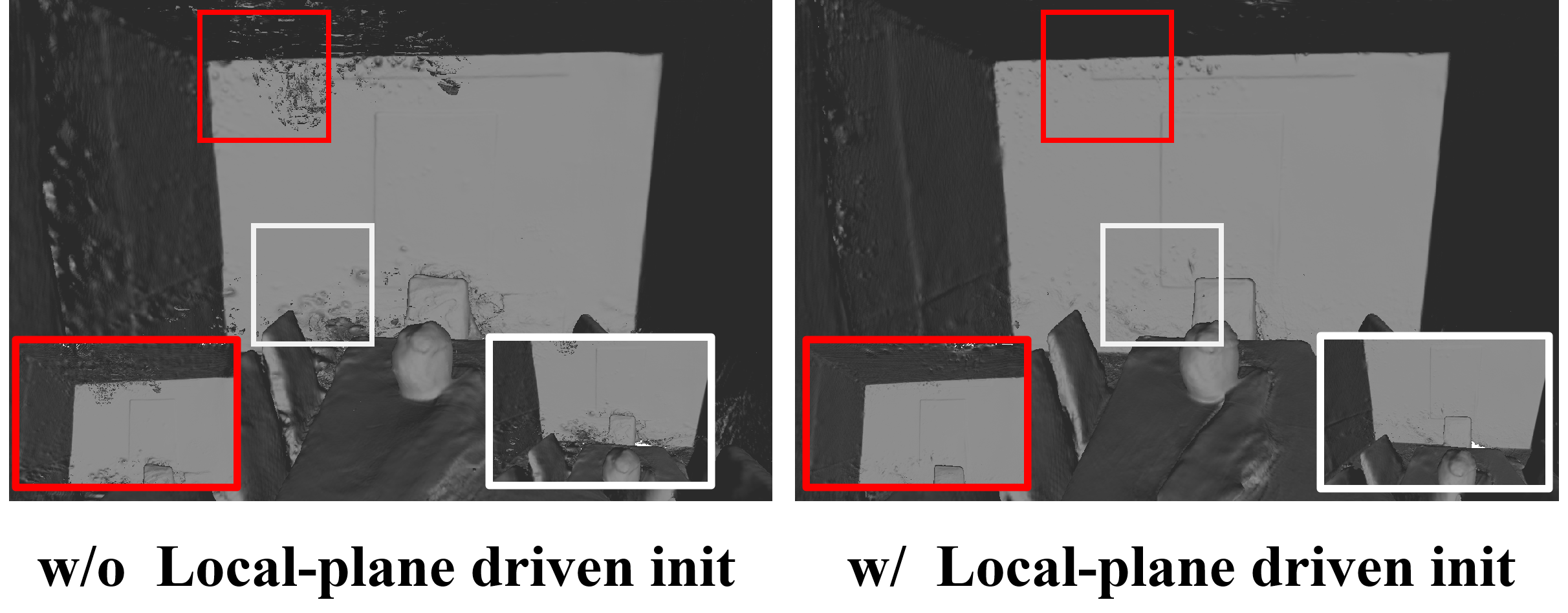}
\caption{Qualitative ablation of local-plane driven initialization. PCA-aligned primitives (right) converge more rapidly to the correct planar orientation, whereas randomly oriented isotropic primitives (left) fail to represent the surface accurately under the same optimization budget.}
\label{fig:ablation_init}
\end{figure}

\subsection{Geometry Reconstruction Performance}
\label{subsec:recon}
We compare the geometry reconstruction quality of GeoGS-SLAM against representative NeRF-based and GS-based SLAM systems on Replica and ScanNet++. 

As illustrated in Table~\ref{tab:replica_recon}, GeoGS-SLAM achieves better reconstruction quality on the Replica dataset compared with DROID-SLAM\cite{teed2021droid}, NeRF-based methods NICER-SLAM\cite{zhu2024nicer}, GO-SLAM\cite{zhang2023go}, HI-SLAM\cite{zhang2023hi} and GS-based methods Splat-SLAM\cite{sandstrom2025splat}, HI-SLAM2\cite{zhang2025hi}. Methods marked with $*$ are tested in their online modes without offline optimization for map refinement.

Fig.~\ref{fig:replica_qualitative} demonstrates several qualitative results. The reconstructed meshes of GeoGS-SLAM show more accurate surfaces and fewer artifacts than the baseline methods.

Table~\ref{tab:scannetpp_joint} shows a similar trend on the ScanNet++ dataset, where GeoGS-SLAM outperforms HI-SLAM2\cite{zhang2025hi} on the metric of Chamfer Distance. The qualitative results in Fig.~\ref{fig:scannet++_qualitative} and Fig.~\ref{fig:scannet_qualitative} are consistent with the quantitative comparison: GeoGS-SLAM reconstructs cleaner surfaces than the baseline methods.

To further illustrate the robustness to illumination variations, we qualitatively evaluate reconstruction results obtained under different lighting conditions, as shown in Fig.~\ref{fig:illumination}. By focusing exclusively on geometric reconstruction without color modeling, GeoGS-SLAM reduces the sensitivity to illumination variations during mapping.

\subsection{Tracking Performance}
\label{subsec:tracking}
We evaluate online tracking performance on the ScanNet++, ScanNet, and Replica datasets. The quantitative results are reported in Table~\ref{tab:scannetpp_joint}, Table~\ref{tab:scannet_tracking}, and Table~\ref{tab:replica_tracking}.

On the ScanNet++ dataset, GeoGS-SLAM achieves an ATE of 4.41 cm, compared with 4.81 cm of HI-SLAM2\cite{zhang2025hi}. On the ScanNet dataset, GeoGS-SLAM achieves the lowest ATE of 7.00 cm, outperforming all baseline methods. On the Replica dataset, GeoGS-SLAM also achieves the lowest ATE among the compared methods. These results demonstrate that GeoGS-SLAM provides consistently accurate camera tracking across both synthetic and real-world benchmarks.

\begin{figure}[t]
\centering
\includegraphics[width=0.49\textwidth]{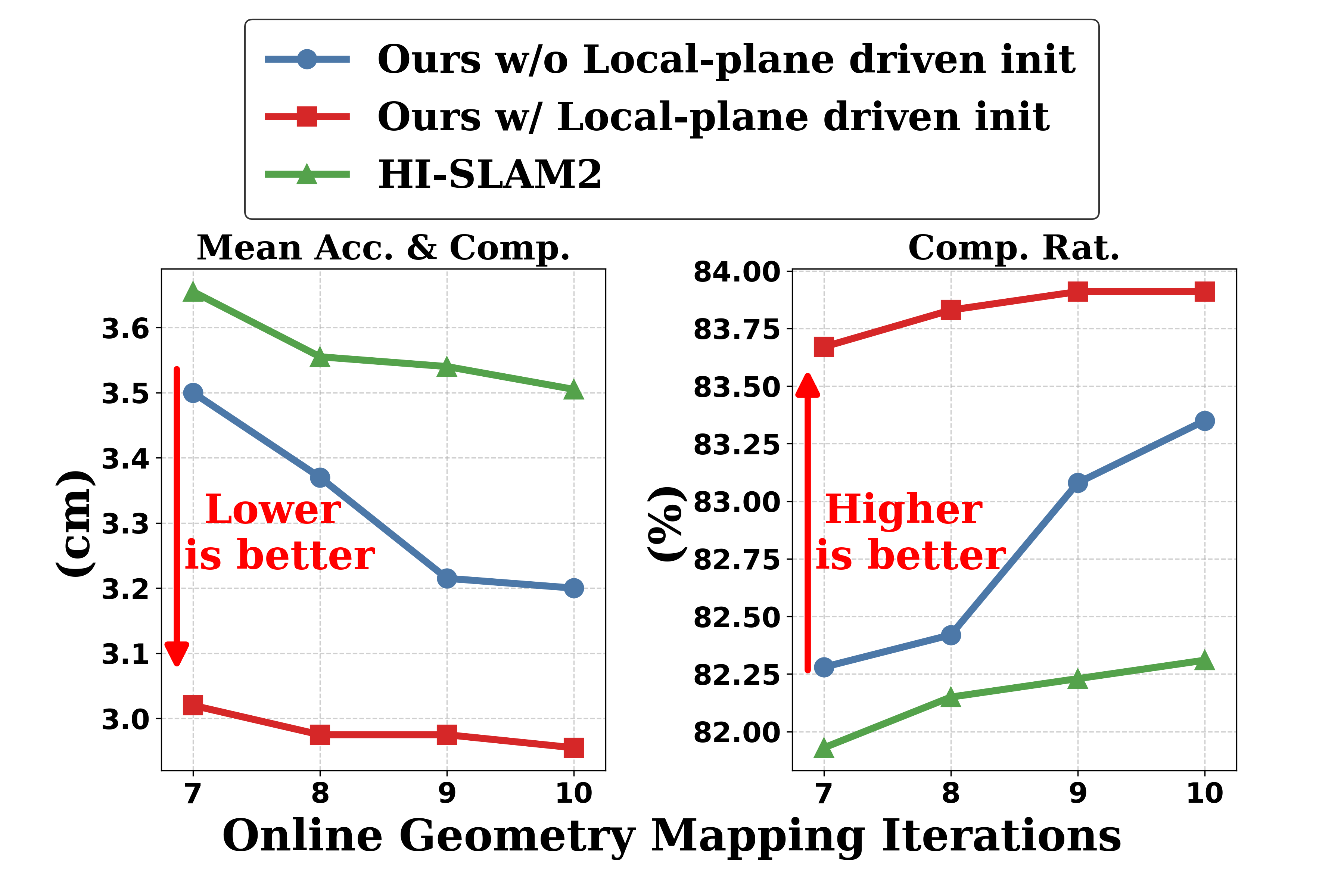}
\caption{
Ablation study of the proposed local-plane driven initialization on the Replica dataset under different geometry mapping iterations varied from 7 to 10. Left: average of mean Accuracy (Acc.) and mean Completeness (Comp.), lower is better. Right: Completeness Ratio (Comp.Rat.), higher is better. The proposed initialization achieves better geometry with fewer iterations.
}
\label{fig:ablation_init_plot}
\end{figure}

\begin{figure}[t]
\centering
\includegraphics[width=0.49\textwidth]{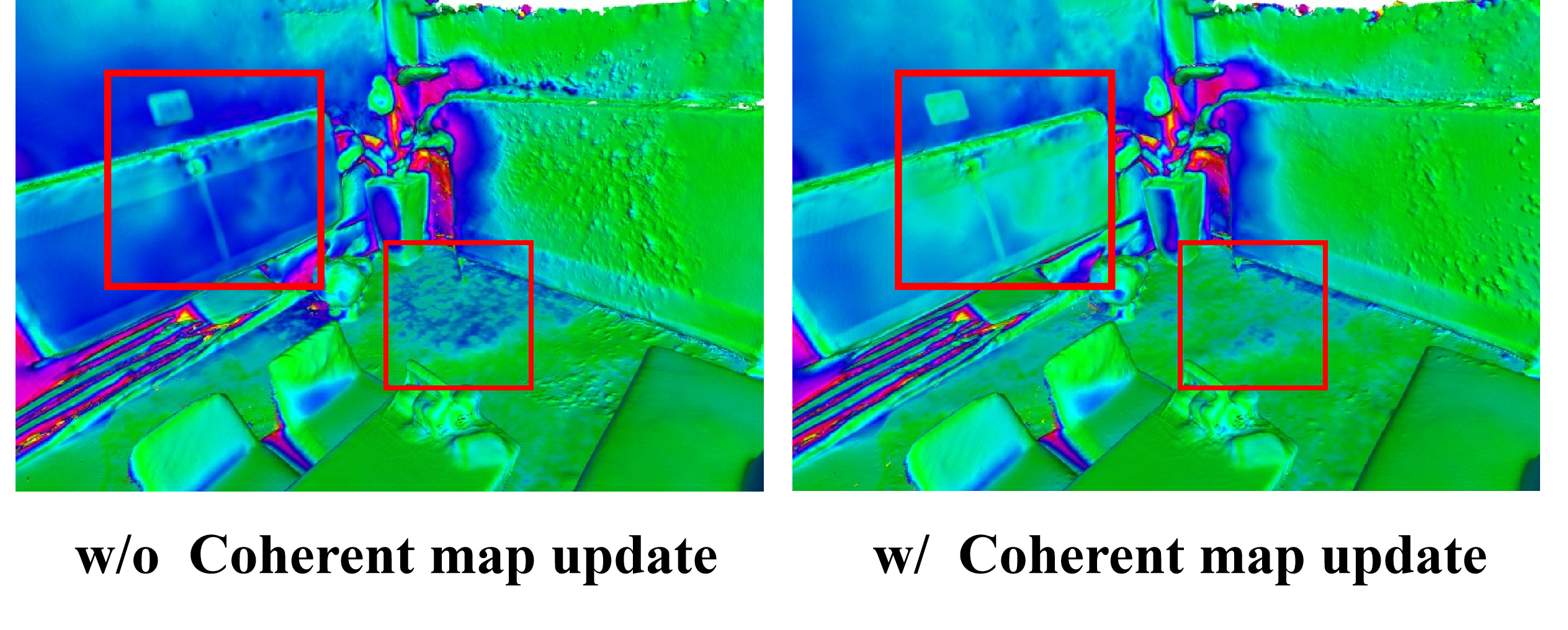}
\caption{Qualitative ablation of the coherent map update. Left: the per-frame propagation update. Right: the proposed coherent update. Green denotes low geometric error, whereas blue, purple, and yellow indicate progressively larger errors. The coherent update better preserves local structural consistency after trajectory correction.}
\label{fig:ablation_map_update}
\end{figure}

\begin{table}[t]
\caption{Ablation on local-plane driven initialization.}
\label{tab:ablation_init}
\centering
\begin{tabular}{|c|c|c|}
\hline
\multirow{3}{*}{w/o local-plane init - 7\,iters} & Acc.$\downarrow$ & 2.77 \\
 & Comp.$\downarrow$ & 4.23 \\
 & Comp.Rat$\uparrow$ & 82.28 \\
\hline
\multirow{3}{*}{w/o local-plane init - 10\,iters} & Acc.$\downarrow$ & 2.37 \\
 & Comp.$\downarrow$ & 4.03 \\
 & Comp.Rat$\uparrow$ & 83.35 \\
\hline
\multirow{3}{*}{\textbf{Ours(w/local-plane init)} - 7\,iters} & Acc.$\downarrow$ & 2.09 \\
 & Comp.$\downarrow$ & 3.95 \\
 & Comp.Rat$\uparrow$ & 83.67 \\
\hline
\multirow{3}{*}{\textbf{Ours(w/local-plane init)} - 10\,iters} & Acc.$\downarrow$ & \textbf{1.98} \\
 & Comp.$\downarrow$ & \textbf{3.93} \\
 & Comp.Rat$\uparrow$ & \textbf{83.91}    \\
\hline
\end{tabular}
\end{table}
\begin{table}[t]
\caption{Ablation on coherent map update strategy.}
\label{tab:ablation_map_update}
\centering
\begin{tabular}{|c|c|c|}
\hline
\multirow{3}{*}{Per-frame propagation update} & Acc.$\downarrow$ & 2.16 \\
 & Comp.$\downarrow$ & 4.05 \\
 & Comp.Rat$\uparrow$ & 83.15 \\
\hline
\multirow{3}{*}{\textbf{Ours(w/global map update)}} & Acc.$\downarrow$ & \textbf{2.09} \\
 & Comp.$\downarrow$ & \textbf{3.95} \\
 & Comp.Rat$\uparrow$ & \textbf{83.67} \\
\hline
\end{tabular}
\end{table}
\begin{table}[t]
\caption{Sensitivity to the $\texttt{pose\_lr}$ on the Replica dataset. ATE is reported in cm$\downarrow$.}
\label{tab:ablation_pose_lr}
\centering
\begin{tabular}{|c|c|}
\hline
$\text{pose\_lr}$ & ATE $\downarrow$ \\
\hline
$1\times10^{-3}$ & 0.29 \\
$6\times10^{-4}$ & \textbf{0.28} \\
$4\times10^{-4}$ & \textbf{0.28} \\
$2\times10^{-4}$ & \textbf{0.28} \\
$1\times10^{-4}$ & 0.29 \\
\hline
\end{tabular}
\end{table}

\subsection{GeoGS Reconstruction Evaluation}
\label{subsec:rep_val}
Motivated by the limited optimization time available in SLAM mapping, we evaluate all methods under the same time-constrained setting, with optimization limited to 2 minutes for each method.

Following representative GS-based surface reconstruction methods, we evaluate GeoGS on 15 DTU sequences against 2DGS\cite{huang20242d}, PGSR\cite{chen2024pgsr}, and QGS\cite{zhang2025quadratic}. As illustrated in Table~\ref{tab:dtu_main}, GeoGS achieves a mean Chamfer distance of 0.62\,cm, significantly outperforming 2DGS (1.38\,cm), PGSR (2.07\,cm), and QGS (1.73\,cm). The advantage is consistent across most sequences with varying geometry complexity, including smooth curved surfaces and thin structures.

We further test the convergence in geometric accuracy of GeoGS. As shown in Fig.~\ref{fig:dtu55-convergence}, the qualitative results shows cleaner and more complete surfaces with much faster convergence. Fig.~\ref{fig:plot-convergence-dtu} further shows that the convergence curve of GeoGS decreases more steeply during early optimization. We further evaluate the geometric convergence on the Replica Office0 sequence. As shown in Fig.~\ref{fig:replica-office0-convergence}, GeoGS achieves faster geometric convergence with smoother surfaces.

Beyond the superior geometric convergence speed discussed above, we report a quantitative comparison on the DTU Scan55 sequence to evaluate the model storage and primitive efficiency. As illustrated in Table~\ref{tab:param_eff}, GeoGS requires significantly fewer primitives to represent the scene geometry with 23k Gaussians, which is only 14.3\%, 10.0\%, and 11.5\% of the primitive counts required by 2DGS, PGSR, and QGS, respectively. Furthermore, by removing color-related parameters within our GeoGS primitives, the storage size per primitive is substantially reduced. Consequently, the total model storage drops to 1.2 MB, which shows a significant reduction compared to 2DGS (37.5 MB, 3.2\%), PGSR (54.6 MB, 2.2\%), and QGS (8.1 MB, 14.8\%).

\begin{table}[t]
\caption{Loss-term contribution on the DTU dataset(2-minute budget). CD is reported in cm$\downarrow$.}
\label{tab:ablation_loss}
\centering
\begin{tabular}{|c|c|}
\hline
Configuration & CD (cm) $\downarrow$ \\
\hline

w/o $\mathcal{L}_{\mathrm{mv}}$ & 9.04 \\
w/o $\mathcal{L}_{\mathrm{nc}}$ & 0.67 \\
w/o $\mathcal{L}_{\mathrm{ds}}$ & 0.56 \\
\hline
\textbf{GeoGS} & \textbf{0.53} \\
\hline
\end{tabular}
\end{table}

\subsection{Ablation Studies}
\label{subsec:ablation}
We conduct ablation studies to isolate the contribution of different components in GeoGS-SLAM: local-plane driven initialization, Coherent map update strategy, loss-term contribution in the training framework and the learning rate of pose in online pose correction.

\textbf{local-plane driven initialization.}
We ablate local-plane driven initialization strategy on the Replica dataset. As reported in Table~\ref{tab:ablation_init}, the proposed initialization strategy improves the quality of geometric reconstruction. As shown in Fig.~\ref{fig:ablation_init}, with our local-plane driven initialization strategy, Gaussian primitives align with the estimated local surface geometry from the start, providing a more accurate geometric state for subsequent optimization, leading to fewer artifacts and cleaner surfaces. The convergence curves in Fig.~\ref{fig:ablation_init_plot} further demonstrate that the proposed strategy achieves higher reconstruction quality with fewer optimization iterations.

\textbf{Coherent map update.}
We ablate the proposed coherent map update strategy on the Replica dataset, where loop closure is disabled and the only pose correction is a final global BA followed by map update at the end of the online stage. To isolate the effect of the update strategy from tracking, we save the Gaussian map before global BA and apply both update strategies to the same Gaussian map: per-frame propagation update and our coherent update. As reported in Table~\ref{tab:ablation_map_update}, the coherent update improves the reconstruction metrics. Qualitative results in Fig.~\ref{fig:ablation_map_update} further show that our map update strategy alleviates local structral tearing and reduces artifacts.

\textbf{Pose-step learning rate.}
We ablate the sensitivity of tracking to $\text{pose\_lr}$, the learning rate of pose, on the Replica dataset. As shown in Table~\ref{tab:ablation_pose_lr}, pose refinement improves tracking accuracy when an appropriate learning rate is used.

\textbf{Loss-term contribution.}
We ablate the contribution of individual loss terms by removing each term from the full model and measuring the Chamfer Distance (CD) on the DTU Scan24 sequence under a 2-minute optimization budget without optional geometric priors. The results are reported in Table~\ref{tab:ablation_loss}. Removing any loss term leads to performance degradation, demonstrating that each component contributes to the final reconstruction quality. Among all terms, removing $\mathcal{L}_{\mathrm{mv}}$ results in the largest increase in CD, indicating its critical role in maintaining multi-view geometric consistency. Removing $\mathcal{L}_{\mathrm{nc}}$ or $\mathcal{L}_{\mathrm{ds}}$ also degrades geometric reconstruction quality, confirming their complementary contributions.

\section{Conclusion}
In this paper, we propose Geometry-only Gaussian Splatting (GeoGS), a novel 3DGS-based visual reconstruction approach that directly reconstructs scene geometry. By retaining only geometry-related parameters, GeoGS significantly reduces the complexity of Gaussian primitives. Moreover, without modeling scene appearance, GeoGS requires fewer primitives to represent the scene geometry and is inherently less sensitive to illumination variations. Together, these advantages enable faster geometric convergence. We further develop GeoGS-SLAM, a dense visual SLAM system built upon GeoGS.

We present an effective framework that directly reconstructs scene geometry without color modeling. A local-plane driven initialization strategy on Gaussian primitives is also introduced to accelerate geometric convergence under limited per-keyframe updates budget. Furthermore, to preserve global consistency after trajectory correction in loop closure or global BA, we develop a coherent map update strategy that estimates a unified $\mathrm{Sim}(3)$ transformation over revisited Gaussians.

Extensive experiments on the Replica, ScanNet++, and ScanNet datasets demonstrate that GeoGS-SLAM consistently improves reconstruction quality and tracking accuracy over existing GS-based SLAM systems. In addition, experiments on the DTU dataset further validate the effectiveness of the proposed GeoGS for geometry reconstruction. Together, these results demonstrate the effectiveness of our method for both dense visual SLAM and geometry reconstruction.

\bibliographystyle{IEEEtran}
\bibliography{main}

\end{document}